\documentclass[11pt]{article}

\usepackage[utf8]{inputenc}
\usepackage[T1]{fontenc}
\usepackage{times}
\usepackage{microtype}
\usepackage{amsmath,amssymb}
\usepackage{graphicx}
\usepackage{booktabs}
\usepackage{multirow}
\usepackage{xcolor}
\usepackage{hyperref}
\usepackage{cleveref}
\usepackage{enumitem}
\usepackage{algorithm}
\usepackage{algorithmic}
\usepackage{caption}
\usepackage{subcaption}
\usepackage[margin=1in]{geometry}
\usepackage{authblk}
\usepackage{colortbl}
\usepackage{tikz}
\usepackage{pgfplots}
\pgfplotsset{compat=1.18}
\usetikzlibrary{arrows.meta, positioning, fit, backgrounds, calc}

\definecolor{novelblue}{RGB}{0,102,204}
\definecolor{lightgray}{RGB}{240,240,240}
\definecolor{tierline}{RGB}{180,180,180}
\definecolor{todocolor}{RGB}{200,0,0}

\definecolor{tealfill}{RGB}{225,245,238}
\definecolor{tealstroke}{RGB}{15,110,86}
\definecolor{purplefill}{RGB}{238,237,254}
\definecolor{purplestroke}{RGB}{83,74,183}
\definecolor{coralfill}{RGB}{250,236,231}
\definecolor{coralstroke}{RGB}{153,60,29}
\definecolor{bluefill}{RGB}{230,241,251}
\definecolor{bluestroke}{RGB}{24,95,165}
\definecolor{greenfill}{RGB}{234,243,222}
\definecolor{greenstroke}{RGB}{59,109,17}
\definecolor{amberfill}{RGB}{250,238,218}
\definecolor{amberstroke}{RGB}{133,79,11}
\definecolor{grayfill}{RGB}{241,239,232}
\definecolor{graystroke}{RGB}{95,94,90}
\definecolor{annotgray}{RGB}{120,120,115}
\definecolor{smegreen}{RGB}{76,153,0}
\definecolor{sonettorange}{RGB}{230,159,0}
\definecolor{ensred}{RGB}{204,60,60}
\definecolor{reconcpurple}{RGB}{117,80,178}
\definecolor{fullpipe}{RGB}{0,114,178}
\definecolor{tierline}{RGB}{120,120,120}

\hypersetup{
    colorlinks=true,
    linkcolor=black,
    citecolor=blue,
    urlcolor=blue,
}

\title{Evidence-Grounded Ensemble Diagnosis of 802.11 Packet Captures:\\
A Multi-Stage Pipeline with Deterministic Reliability Scoring}

\author{Jerome Henry}
\author{Swadhin Pradhan}
\author{Miroslav Popovic}
\affil{Cisco Systems}

\date{}

\begin{document}
\maketitle

\begin{abstract}
Diagnosing connectivity problems from 802.11 packet captures necessitates expert-level knowledge of protocols, is a slow process, varies in consistency among engineers, and lacks scalability. Recent approaches utilizing large language models (LLMs) yield analyses that sound plausible but exhibit three significant failures: they fabricate protocol events that are not present in the capture (particularly in truncated traces), their self-reported confidence levels are not calibrated, and their evaluation against human-annotated golden references is biased towards the model that assisted in creating the reference.

We introduce \textsc{PROBE} (Protocol Reasoning Over evidence-Based Ensembles), a multi-stage diagnostic pipeline designed to rectify all three failure modes. This system integrates (i) a deterministic PCAP-to-text normalization process that maintains frame-level verifiability, (ii) a multi-run, multi-candidate ensemble that includes an optional cross-model second opinion and progressive obfuscation, (iii) a verdict-aware evidence framework that considers the absence of failure evidence as contributing evidence, and (iv) a fully deterministic composite reliability score derived from evidence validity, run-to-run stability, and cross-model agreement—without depending on LLM self-assessment.

In a study involving 87 enterprise Wi-Fi captures (104 capture-reviewer pairs), we observe that single-pass LLM analysis enhances the weighted evidence $F_1$ from 0.871 (the human expert baseline) to 0.912, yet it fails to identify diagnostically critical frames in 35\% of instances. Naive ensemble voting results in a performance drop below the expert baseline (0.842), as majority voting tends to amplify conservative verdicts: 50\% of confirmed failures are incorrectly classified as 'no issue' or 'insufficient evidence.' Incorporating a reconciliation step that assesses all candidates against the packet evidence elevates performance to 0.957, achieving a 96\% auto-accept rate and a worst-case floor exceeding 0.70. LLM self-reported confidence consistently clusters at 0.95, irrespective of diagnostic difficulty (71\% of cases report precisely 0.95), indicating that it is uninformative. We further introduce a
model-agnostic evaluation framework based on per-field assertion
matching, which eliminates the circular bias inherent in golden
references co-produced by a specific model.
\end{abstract}

\section{Introduction}
\label{sec:intro}

Enterprise Wi-Fi troubleshooting from packet captures (PCAPs)
requires deep protocol expertise.  A single 802.11 session capture may
contain dozens of relevant management, control, and data frames whose
interpretation depends on subtle ordering, timing, status codes, and
cross-frame state machines.  Subject-matter experts (SMEs) who can
reliably diagnose such captures are scarce, careful and therefore slow and expensive when efficient, and unfortunately sometimes inconsistent with one another.

Large language models offer a path toward fast and automated PCAP diagnosis:
given a textual representation of the capture, they can produce
structured explanations that identify protocol phases, highlight
anomalous frames, and propose root causes.  However, single-pass LLM
analysis exhibits three specific, measurable failure modes:
\begin{enumerate}[leftmargin=*,nosep]
\item \textbf{Hallucinated completion.}  When a capture is truncated
  (e.g., ends mid--four-way handshake), models routinely infer a
  failure that is not evidenced by the packets present. When a packet is missing in an exchange, models often imagine that packet because the next one is present.
\item \textbf{Uncalibrated confidence.}  When asked for its confidence on a self-produced diagnosis, models excel at quantified hand waving, routinely proposing confidence
  scores cluster at 0.85--0.95 regardless of actual diagnostic
  difficulty (\Cref{sec:calibration}).
\item \textbf{Golden reference bias.}  It is tempting to combine SME and LLM diagnoses to produce a Golden diagnosis. However, this structure alone tends to combine weaknesses more than produce strength. Additionally, switching to a different
  model artificially penalizes stylistic divergence rather than diagnostic error
  (\Cref{sec:eval-framework}).
\end{enumerate}

This paper makes three contributions:

\begin{enumerate}[leftmargin=*,nosep]
\item A \textbf{multi-stage ensemble pipeline} that generates
  $N\!\times\!M$ candidate diagnoses (across $N$~runs and
  $M$~candidates per run), with optional cross-model second opinion,
  progressive obfuscation, and a formal verdict taxonomy that includes
  \texttt{INSUFFICIENT\_EVIDENCE} (\Cref{sec:system}). We show that such ensemble increases the reliability of the diagnosis, but only when used with parsimony.
\item A \textbf{deterministic composite reliability score} computed
  from evidence validity, verdict stability, and cross-model
  agreement, without relying on LLM self-assessment, that enables
  principled confidence-based escalation to human review
  (\Cref{sec:reliability}).
\item A \textbf{model-agnostic evaluation framework} based on
  per-field assertion matching that eliminates circular bias in golden
  references (\Cref{sec:eval-framework}).
\end{enumerate}

\section{Background and Motivation}
\label{sec:background}

\subsection{PCAP Diagnosis as a Structured Reasoning Task}
\label{sec:bg-structured}

Unlike many LLM evaluation domains where ground truth is uncertain
(\emph{e.g.,} medical diagnosis) or inherently subjective (\emph{e.g.,} legal reasoning), a
packet capture is a complete, deterministic record of protocol
events.  Frame~$n$ either contains an EAPOL Message~3 or it does not;
the RSSI was either $-79$\,dBm or it was not.  This property creates
a unique opportunity for \emph{evidence-grounded evaluation} that is
stronger than what is possible in most diagnostic settings. The packet capture of an event should contain specific frames, and specific fields of these frames, that explained what took place. 

In the real world, the capture of the exchanges leading to the failure does not always contain the evidence that a user-support analyst dreams of:
\begin{enumerate}
\item Some captures contain direct error messages that seem to point to a deterministic root cause (\emph{e.g.,} the RADIUS server rejected the association because the password provided was invalid). Unfortunately, roots often come in different depths (is the password incorrectly configured on the client or on the server?), and obvious error messages do not always lend themselves to an articulable definitive root cause diagnosis and its associated remedy.
\item Some captures contain indirect evidence of issues (\emph{e.g.,} timeout error: the RADIUS server did not respond in time). Without an additional message collected from another location (a network node on the path, the server logs), it is difficult to make a conclusion similar to 1. Quantitative data or experienced troubleshooters may connect the two categories (\emph{e.g.,} the culprit is likely router 5, because it was faulty the last 27 times this happened).
\item Some captures do not contain any clear message or hint (\emph{e.g.,} the server response is not in the capture, because the capture is truncated, because the capturing device did not record the server response, or another unknown reason), and there is not enough evidence to make any conclusion. Additional information is necessary to bring the issue back to category 2 or 1.  
\end{enumerate}

Captures of type 1 are ideal, but experienced (human) troubleshooters know that network captures are frail traces of events. The view from the capturing device is different from the views from the source and destination of the packet; timing, buffer or driver glitches may prevent the capture to show a frame that the intended target did receive. An error message visible in the capture may not be the revealing clue about the investigated issue, but may merely be an accessory symptom of a minor event. The expert troubleshooter navigates these uncertainties, filling gaps when possible and questioning the capture when needed. Delegating the troubleshooting ask to an LLM supposes that the model learns the same navigation skills.

\subsection{Why Single-Pass and Multi-pass LLM Analysis Fail}
\label{sec:bg-fails}

Delegating the packet capture analysis to an LLM, even when it is fine-tuned, often proves disappointing, because the primary purpose of the model is to produce probabilistically viable tokens, not to apply analysis rigor.  Three concrete examples from our dataset illustrate this challenge and motivate the
pipeline design in this paper.

\paragraph{Example~1: Hallucinated handshake failure.}
An automated script captured a client association. The 802.11 client exchanged discovery messages (probes) with the Access Point, then proceeded through the 802.11 authentication and association phases. The AP then sent the first message (M1) of the four-way handshake, to which the client responded with the expected M2. The capture was interrupted at that point. The human expert rightfully noted that the capture was inconclusive, while Sonnet 4.5 concluded (likely also noting that the captures were intended to troubleshoot network issues) that the fact that the AP did not send M3 indicated a rejection of the client because of an incorrect passphrase. Such false positive would trigger unnecessary troubleshooting of a credential issue that may not exist. This type of issue motivates the INSUFFICIENT EVIDENCE verdict and verdict-aware evidence rules proposed in this paper.

\paragraph{Example~2: Vague SME annotation.}
In another case, the third message (M3) of the 4-way handshake was missing from the capture, but the 4th message (M4) was present, indicating that the exchange completed successfully (the capturing device likely failed to capture M3). The human expert noted in passing the missing frame. Sonnet 4.5, in a first iteration, described the M3 frame, asserting that it was present. In a second iteration, the same model noted that the frame was missing and concluded that the 4-way handshake failed. This issue, and the difference between two iterations of the same model on the same capture, underline that the limitations of the captures do not constrain the model, they enable it to fill in whatever sounds plausible. This type of issues motivates the reconciler's role that this paper suggests in comparing LLM claims against PCAP evidence.

\paragraph{Example~3: Manufactured issue.}
In another case, a client attempts a DNS resolution, first using secure DNS (port 853 over TLS) then, as the server did not respond on the secured port, using regular DNS on port 53. The client then successfully obtains the IP address of the queried URL. Yet Sonnet 4.5 concludes that the DNS resolution failed, because the secured query was not successful. The model cited real frames with real protocol events but drew a diagnostic conclusion that isn't supported: the observation is correct but the inference is wrong. This motivates the "contributing vs non-contributing evidence" distinction and the ensemble's ability to surface disagreement that this paper suggests.

\section{Related Work}
\label{sec:related}

PROBE is at the intersection of four active research areas:
LLM-based network analysis, self-consistency and ensemble reasoning,
LLM evaluation for diagnostic tasks, and evidence-grounded output
assessment.  We review each thread and identify the specific gaps that
PROBE addresses.

\subsection{LLM-Based Network Analysis and Troubleshooting}
\label{sec:rw-network}

The utilization of language models in network data has advanced rapidly since 2024, encompassing packet-level analysis, configuration synthesis, and incident diagnosis.

\paragraph{Packet capture analysis.}
LLMcap~\cite{tulczyjew2024llmcap} applies masked language modeling
(using DistilBERT) to PCAP files for self-supervised failure
detection.  By tokenizing packet headers and training the model to
reconstruct masked fields, LLMcap identifies anomalous packets
through high reconstruction error.  While effective for binary anomaly
detection, LLMcap produces no diagnostic explanation: it flags
\emph{which} packets are anomalous but not \emph{why} or \emph{what
protocol failure} they indicate.  PROBE addresses a fundamentally
different task, producing structured, frame-referenced diagnostic
reasoning rather than binary classification.

Abkenar~\cite{abkenar2025wifi} fine-tunes both encoder-only
(DistilBERT) and decoder-only LLMs for detecting pathologies in IEEE
802.11 networks, including contention, frame loss, hidden terminal
effects, and interference.  The approach achieves high classification
accuracy on supervised data but, like LLMcap, operates at the
pathology-category level without structured evidence or explanatory
reasoning.  PROBE differs in producing per-frame evidence with
contributing/non-contributing annotations and an explicit verdict
taxonomy that includes abstention (\texttt{INSUFFICIENT\_EVIDENCE}).

PLUME~\cite{pradhan2026plume} builds a protocol-native foundation
model for wireless traces, introducing protocol-aware tokenization
that preserves the hierarchical structure of 802.11 frames.  PLUME
operates at a lower abstraction level than PROBE: it learns
representations of packet sequences that can be fine-tuned for
downstream tasks (anomaly detection, traffic classification), while
PROBE operates on textualized PCAP representations and focuses on
diagnostic reasoning and reliability assessment.  The two approaches
are complementary: PLUME's representations could serve as inputs to
PROBE's ensemble pipeline.

\paragraph{Network troubleshooting and diagnosis.}
NetAssistant~\cite{wang2024netassistant} is a dialogue-based network
diagnosis system deployed in ByteDance's data centers for over three
years.  It accepts natural language queries and executes diagnosis
workflows, significantly reducing human oncall burden.  However,
NetAssistant operates through predefined diagnosis workflows rather
than open-ended reasoning over raw packet data, and it does not
address the reliability or consistency of its diagnostic outputs.

BiAn~\cite{tian2025bian} presents an LLM-based framework for failure
localization in Alibaba Cloud's production networks, processing
monitoring data to generate error device rankings with explanations.
BiAn introduces hierarchical reasoning for large-scale data and
prompt refinement through operational feedback.  While BiAn addresses
production-scale diagnosis, it targets device-level fault localization
from aggregated monitoring logs rather than protocol-level diagnosis
from individual packet captures.  PROBE focuses on the complementary
problem of explaining \emph{why} a specific session failed at the
protocol level.

\paragraph{Network-specific LLMs.}
Mobile-LLaMA~\cite{kan2024mobilellama} instruction-fine-tunes
LLaMA~2~13B on 5G network analysis data, demonstrating that
domain-specific fine-tuning improves network data analytics tasks.
NetLLM~\cite{wu2024netllm} adapts general-purpose LLMs for
networking tasks including viewport prediction and adaptive bitrate
streaming.  Both approaches focus on adapting LLM capabilities to
network data but do not address the reliability or consistency of
diagnostic outputs, the central concern of PROBE.

\paragraph{Benchmarking.}
NIKA~\cite{wang2025nika} provides the largest public benchmark for
LLM-driven network incident diagnosis, comprising hundreds of curated
incidents across five network scenarios.  Its evaluation reveals a
critical finding that motivates PROBE: while larger models succeed
more often in \emph{detecting} network issues, they still struggle to
\emph{localize faults and identify root causes}.  PROBE directly
addresses this gap through multi-hypothesis ensemble reasoning and
reconciliation against packet-level evidence, targeting exactly the
root-cause identification task where single-pass LLMs fail.

\subsection{Self-Consistency and Ensemble Reasoning}
\label{sec:rw-consistency}

PROBE's multi-run, multi-candidate ensemble architecture builds on the
self-consistency idea introduced by Wang
et~al.~\cite{wang2023selfconsistency}, which samples diverse
reasoning paths and selects the most consistent answer through
majority voting.  Self-consistency achieves significant improvements
on arithmetic and commonsense reasoning benchmarks (up to +17.9\%
accuracy on GSM8K with PaLM-540B) by leveraging the intuition that
complex problems admit multiple valid reasoning paths to the same
answer.

PROBE extends self-consistency in three important ways that address
the limitations of naive majority voting in diagnostic contexts:

First, \textbf{majority voting fails on diagnostic tasks with
conservative-verdict bias}.  In our experiments
(\Cref{sec:ablation}), a $3\!\times\!3$ ensemble with majority voting
\emph{degrades} performance below the single-pass baseline (Wt~$F_1$
from 0.912 to 0.842), because multiple candidates converge on
conservative verdicts (\texttt{NO\_ISSUE\_FOUND},
\texttt{INSUFFICIENT\_EVIDENCE}) even when a minority candidate
correctly identifies the failure.  Self-consistency assumes a unique
correct answer that the majority will find; diagnostic reasoning has a
systematic bias toward ``no issue'' that violates this assumption.

Second, \textbf{PROBE adds a reconciliation step that replaces
majority voting with evidence-based selection}.  Instead of counting
votes, the reconciler evaluates all candidates against the actual
packet evidence and optional SME annotations, recovering correct
minority diagnoses that majority voting would discard.  This raises
Wt~$F_1$ from 0.842 (majority-voted ensemble) to 0.957 (reconciled
ensemble).

Third, \textbf{PROBE introduces cross-model diversity} through a
second opinion from a different model family (e.g., Llama~3.3~70B
alongside Claude~Sonnet).  Standard self-consistency samples from the
same model, which can produce correlated errors.  Cross-family
diversity provides an independent signal that the reconciler can
exploit.

Several extensions to self-consistency have been proposed to reduce
computational cost. For example,  ESC~\cite{li2024esc} introduces early stopping
when sufficient agreement is reached, reducing sampling by up to 80\%
on some benchmarks.  DSC~\cite{nair2024dsc} adapts sampling budget
to problem difficulty.  These cost-reduction techniques are
complementary to PROBE and could be applied to reduce the ensemble
budget in production deployments.

Multi-agent debate and verification frameworks
(e.g.,~\cite{liang2023encouraging, du2023improving}) use multiple LLM
agents that iteratively critique and refine each other's outputs.
PROBE's architecture is structurally different: rather than using an iterative
process, it uses a single reconciliation pass over independently
generated candidates.  This design choice is deliberate, iterative
debate can converge on a shared narrative that may not reflect the
packet evidence, while independent generation followed by
evidence-grounded reconciliation preserves hypothesis diversity until
the final selection.

\subsection{LLM Evaluation and LLM-as-Judge}
\label{sec:rw-evaluation}

Evaluating the quality of LLM outputs is also a rapidly evolving field.
The LLM-as-a-Judge concept~\cite{zheng2024judging} uses LLMs to
assess the quality of other LLMs' outputs, replacing expensive human
evaluation.  Comprehensive surveys~\cite{li2024llmjudgesurvey,
xiao2025metajudge} identify significant limitations: position bias
(LLM judges are sensitive to the order of presented options),
self-preference bias (models favor their own outputs), and prompt
sensitivity (small changes in evaluation prompts can flip judgments).

\paragraph{Reference-free vs.\ reference-based evaluation.}
Thakur et~al.~\cite{thakur2025nofree} demonstrate that reference-free
LLM evaluation has inherent biases that limit its usefulness,
particularly self-preference bias where the judge model favors outputs
from its own generative distribution.  Providing human-written
reference answers significantly improves judge agreement with human
annotators.  PROBE's evaluation framework is reference-based: it
scores against a golden reference anchored to verifiable packet
evidence, avoiding the circularity of reference-free assessment.

\paragraph{Position bias.}
Shi et~al.~\cite{shi2024position} conduct a systematic study of
position bias in LLM-as-a-Judge, finding that bias varies
significantly across judges and tasks and is strongly affected by the
quality gap between solutions.  PROBE sidesteps position bias by
design: the reconciler receives all candidates simultaneously (not in
pairwise comparison) and evaluates each against the PCAP ground truth
rather than against other candidates.

\paragraph{Domain-specific evaluation.}
In the medical domain, the CLEVER
framework~\cite{liao2025clever} develops expert-driven evaluation of
clinical LLM outputs and finds that LLM self-evaluation exhibits
systematic biases compared to domain expert assessment.
Yang et~al.~\cite{yang2025medeval} automate expert-level medical
reasoning evaluation, demonstrating that structured evaluation rubrics
aligned with clinical workflows outperform generic quality metrics.
These findings directly inform PROBE's evaluation design: rather than
generic similarity scores, we evaluate per-field against
domain-specific criteria (frame coverage, protocol type agreement,
diagnostic conclusion consistency).

\paragraph{Evaluation of structured outputs.}
Most LLM evaluation work focuses on free-text generation (summaries,
translations, code).  Evaluation of structured multi-field diagnostic
outputs, where different fields require different evaluation
strategies and carry different diagnostic importance, is largely
unaddressed.  PROBE contributes a tiered evaluation framework where
frame-level evidence is evaluated through set comparison (no NLP
needed), protocol types through exact match on constrained
vocabularies, and explanatory text through verdict-aware consistency
checking.

\subsection{Evidence Grounding and Factuality}
\label{sec:rw-grounding}

The FACTS Grounding Leaderboard~\cite{jacovi2025facts} benchmarks
LLMs' ability to generate responses fully grounded in provided context
documents.  FACTS evaluates whether every claim in the response is
supported by the input, using multiple LLM judges to reduce evaluator
bias.  The benchmark has been extended to
FACTS~v2~\cite{jacovi2025factsv2}, updating judge models and
expanding evaluation coverage.

PROBE's evidence validity metric is conceptually similar to FACTS
grounding: it verifies that frame numbers cited by the model actually
exist in the PCAP text and that evidence items are logically aligned
with the declared verdict.  However, PROBE operates on structured
diagnostic output (not free-form text) and introduces a novel
dimension not present in FACTS: \emph{verdict-aware grounding}, where
the meaning of ``supporting evidence'' changes depending on the
diagnostic conclusion.  For a \texttt{CONFIRMED\_ISSUE} verdict,
supporting evidence means frames that exhibit the failure.  For an
\texttt{INSUFFICIENT\_EVIDENCE} verdict, supporting evidence means
frames (or their absence) that demonstrate the capture is incomplete, a form of grounding that generic factuality benchmarks do not
address.

\subsection{Positioning of PROBE}
\label{sec:rw-positioning}

\Cref{tab:related-comparison} summarizes how PROBE relates to prior
work across six dimensions.

\begin{table}[t]
  \centering
  \caption{Comparison of PROBE with related work across key
    dimensions.  \checkmark~=~supported, $\circ$~=~partial,
    ---~=~not addressed.}
  \label{tab:related-comparison}
  \small
  \begin{tabular}{@{}lcccccc@{}}
    \toprule
    \textbf{System}
      & \rotatebox{60}{\parbox{2cm}{\scriptsize Structured\\reasoning}}
      & \rotatebox{60}{\parbox{2cm}{\scriptsize Frame-level\\evidence}}
      & \rotatebox{60}{\parbox{2cm}{\scriptsize Multi-model\\ensemble}}
      & \rotatebox{60}{\parbox{2cm}{\scriptsize Reliability\\scoring}}
      & \rotatebox{60}{\parbox{2cm}{\scriptsize Verdict\\taxonomy}}
      & \rotatebox{60}{\parbox{2cm}{\scriptsize Reconciliation}} \\
    \midrule
    LLMcap~\cite{tulczyjew2024llmcap}
      &---   &---        &---   &---   &---   &---   \\
    WiFi Path.~\cite{abkenar2025wifi}
      &---   &---        &---   &---   &---   &---   \\
    PLUME~\cite{pradhan2026plume}
      &---   & $\circ$    &---   &---   &---   &---   \\
    NetAssistant~\cite{wang2024netassistant}
      & $\circ$ &---      &---   &---   &---   &---   \\
    BiAn~\cite{tian2025bian}
      & \checkmark & $\circ$ &---  &---   &---   &---   \\
    NIKA~\cite{wang2025nika}
      &---   &---        &---   & $\circ$ &---  &---   \\
    Self-Cons.~\cite{wang2023selfconsistency}
      &---   &---       & \checkmark & $\circ$ &--- &--- \\
    FACTS~\cite{jacovi2025facts}
      &---   &---        &---   &---   &---   &---   \\
    \midrule
    \textbf{PROBE} (ours)
      & \checkmark & \checkmark & \checkmark & \checkmark & \checkmark & \checkmark \\
    \bottomrule
  \end{tabular}
\end{table}

To the best of our knowledge, no prior work tackles this problem in a unified way. In particular, existing approaches do not combine multi-hypothesis diagnostic reasoning over structured protocol evidence, deterministic reliability scoring that is independent of the model’s own confidence, and an evidence-grounded reconciliation process that brings together both human expertise and multiple model perspectives.

The closest systems we are aware of—such as BiAn, which focuses on production-scale diagnosis, and NIKA, which benchmarks LLM troubleshooting agents—operate at a higher level of abstraction. They tend to emphasize tasks like device ranking or incident classification, rather than producing the kind of frame-level, evidence-annotated diagnostics that PROBE is designed to generate.

Methodologically, the nearest comparison is self-consistency, which relies on majority voting. However, our experiments show that this approach can actually be counterproductive for diagnostic tasks, where it tends to introduce a bias toward overly conservative conclusions.

\section{System Design}
\label{sec:system}

\begin{figure}[t]
\centering
\resizebox{0.92\linewidth}{!}{%
\begin{tikzpicture}[
    >=Stealth,
    node distance=0.55cm,
    every node/.style={font=\sffamily\small},
    stage/.style={
        rectangle, rounded corners=6pt, minimum width=5.0cm,
        minimum height=1.1cm, align=center, draw, thick
    },
    runbox/.style={
        rectangle, rounded corners=5pt, minimum width=4.2cm,
        minimum height=0.85cm, align=center, draw, semithick,
        fill=purplefill, draw=purplestroke
    },
    smallbox/.style={
        rectangle, rounded corners=5pt, minimum width=3.6cm,
        minimum height=0.85cm, align=center, draw, semithick
    },
    decbox/.style={
        rectangle, rounded corners=5pt, minimum width=2.6cm,
        minimum height=0.8cm, align=center, draw, semithick
    },
    arr/.style={->, thick, gray!70},
    dasharr/.style={->, densely dashed, gray!55, semithick},
    annot/.style={font=\sffamily\scriptsize, text=annotgray},
]

\node[stage, fill=grayfill, draw=graystroke] (pcap)
    {\textbf{PCAP file} \\ {\scriptsize Binary packet capture}};

\node[stage, fill=tealfill, draw=tealstroke, below=of pcap] (norm)
    {\textbf{1.\ PCAP normalization} \\
     {\scriptsize tshark $\to$ PDML $\to$ structured text}};

\node[smallbox, fill=grayfill, draw=graystroke,
      left=2.5cm of norm] (sme)
    {\textbf{SME annotation} \\ {\scriptsize (optional)}};

\node[runbox, below=1.1cm of norm] (run1)
    {Run 1: root cause focus};
\node[runbox, below=0.3cm of run1] (run2)
    {Run 2: protocol sequence};
\node[runbox, below=0.3cm of run2] (run3)
    {Run 3: evidence-first};

\node[annot, right=0.2cm of run1.east, anchor=west]
    {$\to$ $M$ candidates};
\node[annot, right=0.2cm of run2.east, anchor=west]
    {$\to$ $M$ candidates};
\node[annot, right=0.2cm of run3.east, anchor=west]
    {$\to$ $M$ candidates};

\node[annot, below=0.15cm of run3] (nxm)
    {$N \times M$ candidates total};

\node[smallbox, fill=coralfill, draw=coralstroke,
      right=2.4cm of run2] (second)
    {\textbf{Second opinion} \\ {\scriptsize Llama 3.3 70B}};

\node[annot, above=0.4cm of second]
    {\textit{Sonnet 4.5 (draft)}};

\begin{scope}[on background layer]
\node[rectangle, rounded corners=10pt, draw=graystroke,
      densely dashed, semithick, inner sep=12pt,
      fit=(run1)(run2)(run3)(second)(nxm)] (ensbox) {};
\end{scope}

\node[annot, anchor=south west, font=\sffamily\footnotesize]
    at (ensbox.north west) {2.\ Draft ensemble (generation tier)};

\node[stage, fill=bluefill, draw=bluestroke,
      below=0.9cm of ensbox.south] (reconcile)
    {\textbf{3.\ Reconciliation} \\
     {\scriptsize Claude Opus 4.1 (judge)}};

\node[stage, fill=tealfill, draw=tealstroke,
      below=of reconcile, minimum width=5.4cm] (scoring)
    {\textbf{4.\ Reliability scoring} \\
     {\scriptsize $E$ (evidence) + $S$ (stability) + $A$ (agreement)}};

\node[decbox, fill=greenfill, draw=greenstroke,
      below left=0.8cm and 0.8cm of scoring] (accept)
    {\textbf{Auto-accept}};

\node[decbox, fill=amberfill, draw=amberstroke,
      below right=0.8cm and 0.8cm of scoring] (review)
    {\textbf{Human review}};


\draw[arr] (pcap) -- (norm);

\draw[arr] (norm) -- (run1);

\draw[dasharr] (sme.south) -- ++(0,-5.6) -| (reconcile.west);

\draw[arr] (ensbox.south) -- (reconcile);

\draw[dasharr] (second.south) |- ([yshift=2pt]reconcile.east);

\draw[arr] (reconcile) -- (scoring);

\draw[arr] (scoring.south west) ++(0.4,0) -- ++(0,-0.2) -| (accept.north);
\node[annot] at ($(scoring.south west)+(0.0,-0.45)$) {$C \geq \theta$};

\draw[arr] (scoring.south east) ++(-0.4,0) -- ++(0,-0.2) -| (review.north);
\node[annot] at ($(scoring.south east)+(0.0,-0.45)$) {$C < \theta$};

\end{tikzpicture}
}%
\caption{Overview of the PROBE pipeline.  The generation tier
  (dashed box) produces $N \times M$ candidate diagnoses from the
  primary draft model plus an independent second opinion.  The
  reconciliation stage synthesizes the best-supported diagnosis by
  evaluating all candidates against the PCAP evidence and optional SME
  annotation.  Deterministic reliability scoring drives the
  accept/review decision without relying on LLM self-assessment.}
\label{fig:pipeline-overview}
\end{figure}
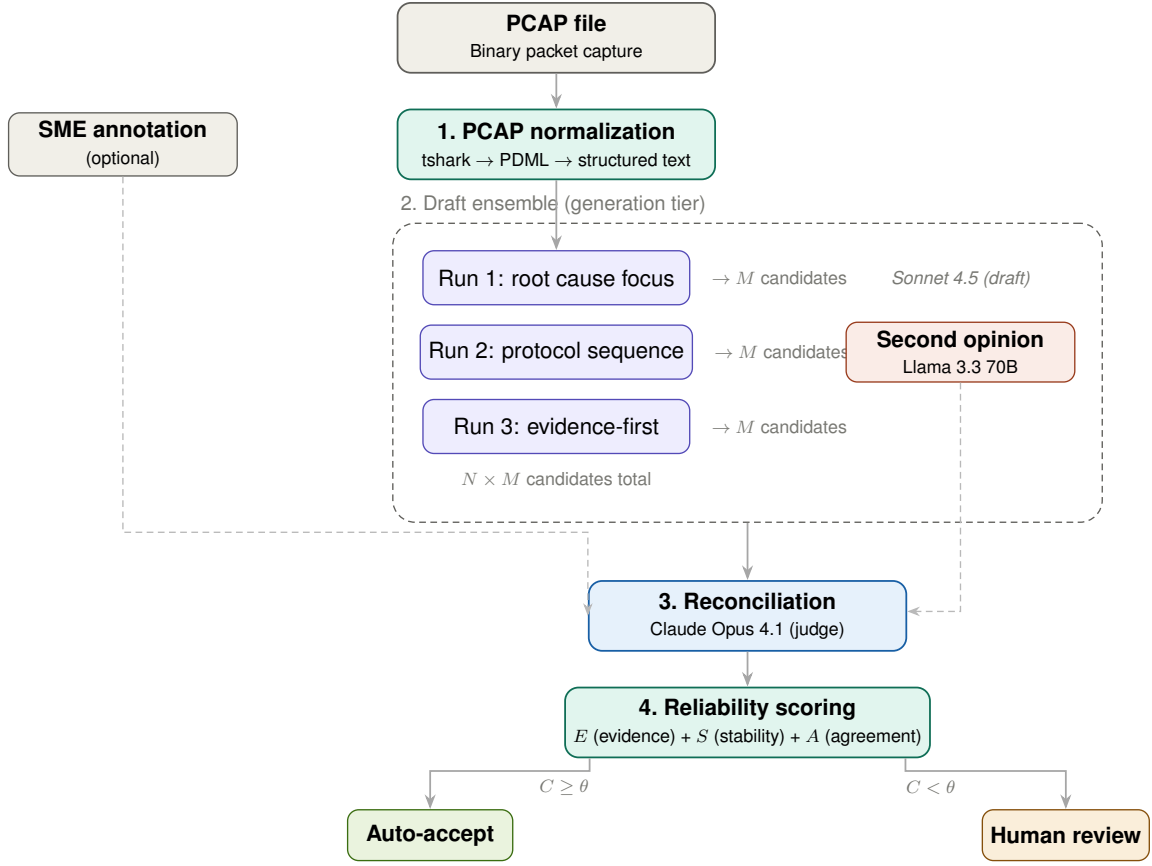

PROBE achieves packet capture root cause analysis through a five-stage
pipeline (\Cref{fig:pipeline-overview}).  Raw captures in PCAP format
are first translated into a semantically meaningful text
representation that preserves frame-level verifiability.  When
available, human expert (SME) annotations are ingested and scored for
quality.  One or more LLMs then independently examine the textualized
capture across multiple runs and produce structured diagnostic
candidates, each containing a frame-by-frame account, a root cause
analysis, evidence annotations, and a formal verdict.  The process
repeats $N$ times with different analytical perspectives, and
optionally with protocol-field obfuscation during later runs to
prevent shallow pattern matching (where the model bases its entire conclusion on a single field with a particular value or message, without considering the full context of the exchange).  An independent second-opinion model
from a different architecture family provides cross-model diversity.
Finally, a reconciliation model reviews all generated opinions
alongside the raw PCAP text and optionally the SME annotations, and anchors its
synthesis to the verifiable packet evidence (existence of pertinent fields and frames, exclusion of non-relevant other messages).

\subsection{PCAP Normalization}
\label{sec:normalization}

The first stage follows PLUME philosophy, and converts a binary PCAP file into a structured textual
representation suitable for LLM consumption.  PROBE uses a script which invokes \texttt{tshark} to produce PDML (Packet Details
Markup Language), then applies a domain-specific textualization layer
that preserves the following information for each frame:

\begin{itemize}[nosep,leftmargin=*]
\item \textbf{Frame number and timestamp.}  Each frame is identified
  by its sequential number in the capture (e.g., ``Frame~120'') and
  its relative timestamp.  Frame numbers serve as the primary evidence
  anchor throughout the pipeline: every claim made by any model must
  reference specific frame numbers (and the relevant fields of interest). The reliability scoring
  verifies that referenced frames actually exist in the capture, and that frames or messages that are indicative or minor or irrelevant issues are not included in the final diagnosis.
\item \textbf{Protocol type and subtype.}  The 802.11 management frame
  type (Probe Request/Response, Authentication, Association,
  Reassociation, Deauthentication, Disassociation), data frame
  classification, and higher-layer protocol (EAPOL, DHCP, DNS, ARP)
  are preserved as structured labels.
\item \textbf{Status and reason codes.}  Authentication and
  association response status codes (e.g., 0x0000 Successful, 0x0011
  AP unable to handle additional STAs), deauthentication reason codes
  (e.g., 0x000f four-way handshake timeout), and EAPOL key
  descriptors are rendered in both hexadecimal and human-readable
  form.
\item \textbf{Radio-frequency metadata.}  Received Signal Strength
  Indicator (RSSI), Signal-to-Noise Ratio (SNR), data rate, channel,
  and retry flag.  These are critical for diagnosing RF-related
  failures (poor signal causing handshake timeout, for example).
\item \textbf{EAPOL handshake state.}  For 802.1X and WPA/WPA2
  sessions, the four-way handshake message number (1/4 through 4/4),
  replay counter, and key information are explicitly labeled.
\item \textbf{Higher-layer details.}  DHCP message types
  (Discover/Offer/Request/Ack/Nack), DNS query and response records,
  and ARP request/reply pairs.
\end{itemize}

The resulting textal representation of the exchange typically ranges from 500 to 5{,}000 tokens
depending on capture length, with each frame occupying 3--15 tokens.
This representation has a critical property: it is \emph{deterministic
and verifiable}.  Given a PCAP file, the textualization always
produces the same output, and every factual claim an LLM makes about
the capture (``frame~126 contains an association response with status
0x0011'') can be mechanically verified against this text.  This useful
property enables the evidence validity component of the reliability
score (\Cref{sec:evidence-validity}).

\subsection{SME Annotation Processing}
\label{sec:sme}

When available, human subject-matter expert annotations are ingested
as a complementary signal.  Each SME annotation consists of three
components:

\begin{itemize}[nosep,leftmargin=*]
\item \textbf{\texttt{pcapSummary}}: A free-text summary of the
  capture, typically 1--3 sentences (e.g., ``STA timed-out while
  performing the 4-way handshake when connecting to Corporate SSID, as it
  stopped responding'').
\item \textbf{\texttt{chainOfThought}}: A narrative trace that SMEs produce in response to a support request.
  The chain of thought part typically walks through the protocol exchange and identifies along the way frames of importance and the failure
  point (e.g., ``AP keeps sending message~1 and incrementing replay
  counter (frames 82--87), while no message~2 from client is
  seen'').
\item \textbf{\texttt{highlightFrames}}: A set of frame numbers the
  SME considers diagnostically relevant, with optional per-frame
  annotations indicating which protocol fields matter (e.g., look at 
  frame~88: \texttt{wlan.fixed.reason\_code}: ``4-way handshake
  timeout (0x000f)'').
\end{itemize}

SME annotations vary significantly in quality.  Our dataset surfaces
three distinct quality tiers:

\paragraph{Strong annotations.}  The SME references specific frames and fields,
names the key protocol events that lead to the problematic part, and articulates a causal chain from observed
behavior to root cause.  These annotations typically contain 5+
highlighted frames and 100+ characters of chain-of-thought.

\paragraph{Weak annotations.}  The SME provides a technically correct
but vague analysis (e.g., ``handshake was interrupted'') with few or no
frame references, and a root cause suggestion that is either insufficiently clear (e.g., ``password was misconfigured'' [where?]) or non-committal (e.g., ``the issue may be the passphrase'').  The pipeline cannot distinguish this from a
speculative diagnosis without additional evidence.

\paragraph{Missing annotations.}  No SME input is available or text is unusable.  In many cases, the chain of thought describes a sequence of succesful events before observing that there is an issue without clear technical qualifiers. The
pipeline must rely entirely on LLM analysis and packet evidence.

PROBE treats the SME annotation as \emph{one opinion among several},
not as ground truth.  The reconciliation stage (\Cref{sec:reconcile})
explicitly compares SME claims against packet evidence and can
override the SME when the PCAP contradicts the annotation.  Our
ablation experiments (\Cref{sec:ablation}) show that SME annotations
achieve the highest key frame recall of any individual signal (0.915), suggesting that SMEs are best at identifying key frames and fields of relevance, when they focus on the task. However, SME annotations also achieve the lowest relevant frame coverage (0.342), confirming that
expert input is valuable but often incomplete.

\subsection{Draft Ensemble Architecture}
\label{sec:ensemble}

The draft stage generates an $N \times M$ candidate matrix by running
the draft model $N$ times, each time requesting $M$ candidate
diagnoses.  This produces $N \times M$ structured diagnostic outputs,
each independently reasoned. The goal of this phase is to avoid the trap of a single LLM shot, where a model ability to perform a root cause analysis on a given capture sample is derived from a single attempt. Even with low temperature, a single attempt may omit key words or focus on secondary aspects that a judge (LLM or human) would classify as being of limited relevance. Just like humans can take a second look with a different perspective, this phase allows the model to make several attempts.

\subsubsection{Run Diversity ($N$ Runs)}

Each of the $N$ runs is an independent call to the draft model
(Claude~Sonnet~4.5 in our experiments) with a different analytical
focus.  Three default focus variants are used:

\begin{enumerate}[nosep,leftmargin=*]
\item \textbf{Root cause and decisive evidence}, ``prefer abstain
  if unproven.''  This variant biases the model toward identifying the
  single most likely root cause and the minimal set of frames that
  prove it, while explicitly permitting abstention.
\item \textbf{Protocol sequence anomalies}, ``authenticate /
  associate / EAPOL / DHCP.''  This variant directs attention to the
  protocol state machine, asking the model to identify where the
  expected sequence deviates from normal behavior.
\item \textbf{Evidence-first}, ``list frames that materially
  contribute to the failure.''  This variant inverts the reasoning
  direction: instead of concluding a root cause and then finding
  evidence, it first identifies anomalous frames and then infers what
  they collectively indicate.
\end{enumerate}

Run diversity tests \emph{reasoning stability}: if the same model
reaches the same conclusion through different analytical lenses, the
diagnosis is more likely to be correct.  Divergence across runs
signals either genuine ambiguity in the capture or a fragile
reasoning chain.

\subsubsection{Candidate Diversity ($M$ Candidates per Run)}

Within each run, the model produces $M$ candidate diagnoses in a
single prompt.  Candidate diversity tests \emph{internal ambiguity}:
when the model surfaces multiple plausible interpretations within a
single reasoning path, the capture likely contains genuinely
ambiguous evidence.

Candidates within a run are less independent than candidates across
runs because they share the same prompt context.  For reliability scoring
purposes (\Cref{sec:stability}), stability is measured across runs
(not within them), making run diversity the stronger signal.

\subsubsection{Candidate Structure}

Each candidate diagnosis is a structured JSON object containing:

\begin{itemize}[nosep,leftmargin=*]
\item \textbf{\texttt{diagnosis}}: A short label (e.g., ``4-way
  handshake timeout due to poor signal'').
\item \textbf{\texttt{verdict}}: One of four values from the strict
  verdict taxonomy (\Cref{sec:verdict-taxonomy}).
\item \textbf{\texttt{pcapSummary}}: A narrative summary of what
  happens in the capture.
\item \textbf{\texttt{chainOfThought}}: Step-by-step reasoning about
  the protocol behavior.
\item \textbf{\texttt{key\_frames}}: Frame numbers that directly
  evidence the root cause.
\item \textbf{\texttt{relevant\_frames}}: Frame numbers that provide
  protocol context (superset of key frames).
\item \textbf{\texttt{evidence}}: A list of evidence items, each
  containing a frame reference, one or more fields, a factual claim, and a boolean
  \texttt{contributes} flag indicating whether the evidence supports
  the stated conclusion.
\item \textbf{\texttt{non\_contributing\_observations}}: Protocol
  events that appear noteworthy but do not materially support the
  diagnosis.  Separating these from contributing evidence prevents the
  common failure mode where the model cites an observation
  (e.g., ``low RSSI'') that \emph{looks} issue-related but does not
  actually contribute to the diagnosed failure.
\item \textbf{\texttt{self\_confidence}}: The model's self-assessed
  confidence (0--1).  As we show in \Cref{sec:calibration}, this
  value is commonly used when using LLM experts. However, it is also uninformative and is retained only for analysis purposes.
\end{itemize}

\subsubsection{Candidate Selection}

From the $N \times M$ candidates examining a given capture, the pipeline selects a single
representative using a deterministic procedure:

\begin{enumerate}[nosep,leftmargin=*]
\item Select the best candidate from each run using an objective scoring function
  that prioritizes (in order): verdict weight (confirmed issues rank
  highest, insufficient evidence lowest), count of contributing
  evidence items, and count of key frames. If a tie occurs, the model asserted self-confidence is used as
  tiebreaker. It is important to note that we do not assign a high value to the self-confidence. Its use is merely that, when two diagnoses are objectively equally good, we pick the one that the model likes most.
\item Among the $N$ per-run selections, group by (verdict, diagnosis
  label) and select the majority group.
\item Within the majority group, choose the candidate with the highest
  evidence score.
\end{enumerate}

This majority-vote selection is the mechanism that, as our experiments
show (\Cref{sec:ablation}), can converge on conservative verdicts
when the ensemble is conflicted.  The reconciliation stage
(\Cref{sec:reconcile}) compensates for this by evaluating all
candidates, including minority diagnoses that majority voting would
discard.

\subsubsection{Verdict Taxonomy}
\label{sec:verdict-taxonomy}

Each candidate must declare a verdict from a strict four-value
taxonomy:

\begin{itemize}[nosep,leftmargin=*]
\item \texttt{CONFIRMED\_ISSUE}: Clear protocol failure with
  frame-level evidence directly establishing the root cause (what \Cref{sec:bg-structured} refers to as type 1 issues).
\item \texttt{PLAUSIBLE\_ISSUE}: A likely failure, but the evidence is
  circumstantial or incomplete (what \Cref{sec:bg-structured} refers to as type 2 issues, e.g., poor signal during handshake but
  no explicit failure frame).
\item \texttt{INSUFFICIENT\_EVIDENCE}: The capture is truncated or
  missing decisive packets.  An issue may or may not exist, but the
  available evidence cannot determine the outcome (what \Cref{sec:bg-structured} refers to as type 3 issues).
\item \texttt{NO\_ISSUE\_FOUND}: The observed protocol exchange
  completes normally with no anomalous behavior.
\end{itemize}

The \texttt{INSUFFICIENT\_EVIDENCE} verdict is a critical design
element.  Without it, models are forced to choose between
\texttt{CONFIRMED\_ISSUE} and \texttt{NO\_ISSUE\_FOUND} on captures
that simply end mid-exchange, leading to the hallucinated
completion problem described in \Cref{sec:bg-fails}.  Allowing
explicit abstention reduces false positives on truncated captures
(\Cref{sec:truncation}).

\subsubsection{Verdict-Aware Evidence Rules}
\label{sec:verdict-aware}

A key innovation of PROBE is that the meaning of ``contributing
evidence'' is defined relative to the verdict:

\paragraph{For \texttt{CONFIRMED\_ISSUE} and
\texttt{PLAUSIBLE\_ISSUE}:}
\texttt{contributes=true} means the evidence materially supports the
claimed protocol failure.  Example: ``Frame~88: deauthentication with
reason code 0x000f (4-way handshake timeout).''

\paragraph{For \texttt{INSUFFICIENT\_EVIDENCE}:}
\texttt{contributes=true} means the evidence supports the conclusion
that the capture is incomplete or that the diagnostic question cannot
be resolved.  Example: ``Capture ends after EAPOL msg~2/4; messages
3/4 and 4/4 are not present.''  Another example: ``No
deauthentication, retry storms, or handshake failure indications
appear before the capture ends.''

\paragraph{For \texttt{NO\_ISSUE\_FOUND}:}
\texttt{contributes=true} means the evidence supports that the
observed exchange completed normally.  Example: ``4-way handshake
completes with message~4/4 in frame~27.''

This verdict-aware interpretation prevents a systematic scoring
artifact: without it, candidates with \texttt{INSUFFICIENT\_EVIDENCE}
verdicts mark all evidence as non-contributing (because nothing
``contributes to a failure''), which drives the evidence validity
component (\Cref{sec:evidence-validity}) toward zero even when the
verdict is correct.  The prompt requires at least two contributing
evidence items for every verdict, ensuring that the reliability score
reflects reasoning consistency rather than verdict category.

\subsubsection{Progressive Obfuscation}
\label{sec:obfuscation}

After an initial set of standard runs, PROBE optionally masks key
protocol fields in the PCAP text before executing additional runs.
Fields that may be masked include RSSI values, specific status/reason
codes, and EAPOL message numbers.

The rationale is adversarial robustness: if a diagnosis survives field
masking, it is grounded in the structural properties of the protocol
exchange (frame ordering, presence/absence of expected messages,
timing patterns), not in a single salient keyword.  If the diagnosis
changes under obfuscation, the model was relying on a shallow cue,
for example keying on ``-99~dBm'' to conclude ``poor signal'' without
checking whether the exchange actually failed.

Obfuscation-induced disagreement feeds the stability component of the
reliability score (\Cref{sec:stability}) and provides additional
diagnostic signal to the reconciler.

\subsection{Second-Opinion Model}
\label{sec:second-opinion}

An optional stage injects cross-model diversity by requesting an
independent analysis from a model of a different architecture family.
In our experiments, we use Llama~3.3~70B (Meta) as the second opinion
alongside Claude~Sonnet~4.5 (Anthropic) as the primary draft model.

The second-opinion model receives the same PCAP text and the same
analytical prompt, but has \emph{no visibility} into the primary
ensemble's candidates.  Its candidates are generated independently and
evaluated through the same structured schema (verdict, evidence, key
frames).

The second opinion serves two purposes:

\begin{enumerate}[nosep,leftmargin=*]
\item \textbf{Cross-model agreement signal.}  If the primary ensemble
  and the second-opinion model converge on the same verdict and key
  frames, then the diagnosis is more likely to reflect genuine evidence
  rather than model-specific reasoning patterns.  This feeds the
  cross-model agreement component (\Cref{sec:agreement}).
\item \textbf{Alternative hypothesis for reconciliation.}  When the
  primary ensemble and the second opinion disagree, the reconciler
  sees both perspectives and can evaluate which is better supported by
  the packet evidence.  This is particularly valuable when the primary
  ensemble's majority vote selects a conservative verdict while the
  second opinion (or a minority primary candidate) correctly
  identifies the issue.
\end{enumerate}

Our ablation experiments (\Cref{sec:ablation}) show that the second
opinion alone (without reconciliation) adds negligible value: Config~E
(ensemble + second opinion, Wt~$F_1$~=~0.845) is statistically
indistinguishable from Config~D (ensemble only, 0.842).  However,
when combined with reconciliation (Config~F, 0.957), the second
opinion provides the reconciler with an independent analytical
perspective that contributes to the pipeline's robustness on edge
cases. In other words, a surface analysis of the second opinion may lead one to discard the step was a waste of time and tokens, but a deeper analysis reveals that the second opinion phase reduces the risk of mis-diagnosis.

\subsection{Reconciliation}
\label{sec:reconcile}

The reconciliation stage is the pipeline component that transforms raw
ensemble diversity into diagnostic quality.  A dedicated
reconciliation model (Claude~Opus~4.1 in our experiments, configured
with extended thinking) receives four inputs:

\begin{enumerate}[nosep,leftmargin=*]
\item The full capture, in the form of the PCAP textual representation, that acts as the deterministic ground
  truth.
\item The SME annotation (when available).
\item The best candidate from each of the $N$ primary ensemble runs,
  plus the best candidate fro the second-opinion selection runs, along with the ensemble's
  computed stability and agreement metrics.
\item The capture metadata (protocol step and reason category, when
  known).
\end{enumerate}

The reconciler produces a structured output with two sections:

\paragraph{Suggested gold.}  A normalized diagnosis containing:
\texttt{pcapSummary} (narrative summary), \texttt{chainOfThought}
(reasoning trace), \texttt{frames.key} (diagnostically critical frame
numbers), \texttt{frames.relevant} (full protocol context frames), and
a confidence score.  This is the pipeline's final diagnostic output.

\paragraph{Structured review.}  An explicit comparison organized as:
\begin{itemize}[nosep,leftmargin=*]
\item \texttt{sme\_claims}: What the SME asserted.
\item \texttt{pcap\_shows}: What the packet evidence actually
  demonstrates.
\item \texttt{therefore}: The logical connection between evidence and
  conclusion.
\item \texttt{changes\_made}: How the suggested gold differs from the
  SME and draft inputs.
\item \texttt{questions\_for\_human}: Open questions that the
  available evidence cannot resolve.
\item \texttt{suggested\_verdict}: APPROVE / REVISE / REJECT /
  UNCERTAIN. The verdict is used to redirect the output (or not) for further human review.
\item \texttt{verdict\_justification}: Reasoning for the suggested
  verdict.
\end{itemize}

The structured review serves two purposes.  For the pipeline, it
provides a traceable audit trail of how the final diagnosis was
derived.  For human reviewers, it focuses attention on specific points
of disagreement rather than requiring a full re-analysis.

\paragraph{What the reconciler does not do.}
The reconciler does \emph{not} determine diagnostic confidence.
Confidence is computed deterministically from the ensemble metrics
(\Cref{sec:composite}).  The reconciler synthesizes the best-supported
diagnosis; the reliability scoring framework independently assesses
how trustworthy that diagnosis is.  This separation prevents the
reconciler from inflating confidence through eloquent justification (a known failure mode of LLM-as-judge approaches where the model's
reasoning quality correlates with its rhetorical persuasiveness rather
than its factual accuracy~\cite{thakur2025nofree}).

\paragraph{Why reconciliation succeeds where majority voting fails.}
Our ablation (\Cref{sec:ablation}) shows that Config~D (majority-voted
ensemble, Wt~$F_1$~=~0.842) is \emph{worse} than single-pass Sonnet
(Config~B, 0.912), while Config~F (same ensemble with reconciliation,
0.957) nearly matches the best configuration.  The mechanism is
clear: majority voting discards minority candidates that may be
correct, because when provided with a choice, models tend to choose conservative options rather than risking a stronger opinion. But then the reconciler evaluates \emph{all} candidates against
the PCAP evidence and can recover correct minority diagnoses.  The
reconciler's access to the raw PCAP text (the deterministic ground
truth) is what makes this possible: it can verify claims directly
rather than relying on vote counts.

\section{Deterministic Reliability Scoring}
\label{sec:reliability}

A key design principle of PROBE is that diagnostic confidence is
\emph{computed from observable signals}, not \emph{elicited from the
model}.  LLM self-reported confidence is known to be poorly calibrated
across domains~\cite{liao2025clever}, and our experiments confirm this
for network diagnosis: 71\% of self-reported draft confidence values
are exactly 0.95 regardless of actual diagnostic difficulty
(\Cref{sec:calibration}).

The composite reliability score $C$ is derived from three
independently measurable signals: evidence validity~$E$, run-to-run
stability~$S$, and cross-model agreement~$A$.  Each signal captures a
different dimension of diagnostic reliability, and all three can be
computed without any additional LLM calls: they are deterministic
functions of the ensemble output and the PCAP text.

\subsection{Evidence Validity ($E$)}
\label{sec:evidence-validity}

Evidence validity measures whether the selected candidate's claims are
grounded in the actual packet capture.  It has two multiplicative
components:

\begin{equation}
  E = E_{\text{frame}} \times E_{\text{contributes}}
\label{eq:evidence-validity}
\end{equation}

The first component, \emph{frame existence}, verifies that cited frame
numbers actually appear in the PCAP text:

\begin{equation}
  E_{\text{frame}} =
    \frac{|\mathcal{F}_{\text{cited}} \cap \mathcal{F}_{\text{pcap}}|}
         {|\mathcal{F}_{\text{cited}}|}
\label{eq:evidence-frame}
\end{equation}

where $\mathcal{F}_{\text{cited}}$ is the union of key frames,
relevant frames, and evidence item frame references from the selected
candidate, and $\mathcal{F}_{\text{pcap}}$ is the set of frame
numbers extracted from the normalized PCAP text.  A model that
references frame~200 in a 53-frame capture receives an immediate
penalty.

The second component, \emph{evidence contribution}, measures whether
the candidate's evidence items are logically aligned with its verdict:

\begin{equation}
  E_{\text{contributes}} =
    \frac{|\{e \in \mathcal{E} : e.\mathit{contributes} = \texttt{true}\}|}
         {|\mathcal{E}|}
\label{eq:evidence-contributes}
\end{equation}

where $\mathcal{E}$ is the set of evidence items.  Under the
verdict-aware evidence rules (\Cref{sec:verdict-aware}), a candidate
with verdict \texttt{INSUFFICIENT\_EVIDENCE} can still achieve high
$E_{\text{contributes}}$ by marking truncation-supporting evidence as
contributing.

The multiplicative formulation ensures that both conditions must be
satisfied: a candidate that cites real frames but marks all evidence
as non-contributing scores low (high $E_{\text{frame}}$, low
$E_{\text{contributes}}$), as does a candidate that marks evidence as
contributing but references nonexistent frames.

\subsection{Run-to-Run Stability ($S$)}
\label{sec:stability}

Stability measures whether independent reasoning attempts converge on
the same diagnostic conclusion, and combines two signals: frame-level
agreement and verdict-level agreement.

\paragraph{Frame stability.}  Let $\mathcal{K}_i$ be the set of key
frames identified by the selected candidate from run~$i$.  Frame
stability is the average pairwise Jaccard similarity across all
$\binom{N}{2}$ run pairs:

\begin{equation}
  S_{\text{frames}} = \frac{1}{\binom{N}{2}}
    \sum_{i<j} J(\mathcal{K}_i, \mathcal{K}_j)
    \quad\text{where}\quad
    J(A, B) = \frac{|A \cap B|}{|A \cup B|}
\label{eq:stability-frames}
\end{equation}

$S_{\text{frames}} = 1.0$ means all runs identified exactly the same
key frames; $S_{\text{frames}} = 0.0$ means no two runs share a
single key frame.

\paragraph{Verdict stability.}  Let $v_i$ be the verdict declared by
the selected candidate from run~$i$.  Verdict stability is the
strength of the plurality verdict:

\begin{equation}
  S_{\text{verdict}} = \frac{\max_v |\{i : v_i = v\}|}{N}
\label{eq:stability-verdict}
\end{equation}

$S_{\text{verdict}} = 1.0$ means all runs agree on the same verdict;
$S_{\text{verdict}} = 1/N$ means every run chose a different verdict.

\paragraph{Combined stability.}

\begin{equation}
  S = \alpha \, S_{\text{frames}} + (1 - \alpha) \, S_{\text{verdict}}
\label{eq:stability}
\end{equation}

In the current implementation, $\alpha = 0.5$, weighting frame-level
and verdict-level agreement equally.  The $\alpha$ parameter is
tunable: increasing it emphasizes fine-grained evidence agreement,
while decreasing it emphasizes categorical diagnostic agreement.

\subsection{Cross-Model Agreement ($A$)}
\label{sec:agreement}

When a second-opinion model is available, cross-model agreement
measures whether an independently-reasoned analysis from a different
model family converges on the same diagnosis.  Let $c_{\text{pri}}$ be
the primary ensemble's selected candidate and $c_{\text{sec}}$ be the
second-opinion model's selected candidate.

\begin{equation}
  A = \beta \cdot \mathbb{1}[v_{\text{pri}} = v_{\text{sec}}]
    + (1 - \beta) \cdot J(\mathcal{K}_{\text{pri}},
    \mathcal{K}_{\text{sec}})
\label{eq:agreement}
\end{equation}

where $v_{\text{pri}}, v_{\text{sec}}$ are the verdicts,
$\mathcal{K}_{\text{pri}}, \mathcal{K}_{\text{sec}}$ are the key
frame sets, and $\beta = 0.5$ in the current implementation.

Cross-model agreement provides a qualitatively different signal from
run-to-run stability: stability measures whether the \emph{same}
model is consistent with itself across prompt variants, while
agreement measures whether \emph{different} models (with different
training data, architectures, and reasoning patterns) converge on the
same conclusion.  Disagreement between Claude~Sonnet and Llama~3.3 is
a stronger signal of genuine ambiguity than disagreement between two
Sonnet runs (even when the runs use different prompts), because the models do not share the same systematic
biases.

When no second-opinion model is used, $A$ is excluded from the
composite, and the weights of $E$ and $S$ are renormalized
accordingly.

\subsection{Composite Confidence and Escalation}
\label{sec:composite}

The composite confidence score combines the three signals into a
single value:

\begin{equation}
  C =
  \begin{cases}
    w_E \cdot E + w_S \cdot S + w_A \cdot A
      & \text{if second opinion available} \\[4pt]
    \displaystyle\frac{w_E}{w_E + w_S} \cdot E
      + \frac{w_S}{w_E + w_S} \cdot S
      & \text{otherwise}
  \end{cases}
\label{eq:composite}
\end{equation}

In the current implementation, the weights are $w_E = 0.40$,
$w_S = 0.30$, $w_A = 0.30$ when a second opinion is available, with
the second opinion receiving 25\% of the total weight and the primary
components receiving 75\%:

\begin{equation}
  C = 0.75 \cdot (0.40 \cdot E + 0.30 \cdot S_{\text{frames}}
    + 0.30 \cdot S_{\text{verdict}}) + 0.25 \cdot A
\label{eq:composite-impl}
\end{equation}

The composite score $C \in [0, 1]$ drives the escalation decision:

\begin{itemize}[nosep,leftmargin=*]
\item $C \geq \theta_{\text{accept}}$: Auto-accept the reconciled
  diagnosis.  In our experiments, $\theta_{\text{accept}} = 0.70$.
\item $C < \theta_{\text{accept}}$: Escalate to human review.  The
  low-confidence flag and the specific component(s) that drove it down
  are recorded for the reviewer.
\item \textbf{Veto rule:} If the majority verdict across runs
  disagrees with the reconciler's selected diagnosis, escalate
  regardless of $C$.  This prevents the reconciler from overriding a
  strong ensemble consensus without human oversight.
\end{itemize}

\paragraph{What the composite score does and does not predict.}
Our calibration experiments (\Cref{sec:calibration}) surface an
important nuance.  The composite score does not reliably predict
whether the pipeline's final output is \emph{correct}, because the
reconciler is effective enough to produce correct output in 96\% of
cases regardless of ensemble agreement.  Instead, the composite score
characterizes \emph{capture difficulty}: low-agreement cases identify
captures with genuinely ambiguous evidence (truncated exchanges,
edge-case protocols) that should be forwarded for human review, not for error correction, but for dataset enrichment.  This distinction, quality gating vs.\
difficulty characterization, is discussed in detail in
\Cref{sec:cal-reframe}.

\section{Model-Agnostic Evaluation Framework}
\label{sec:eval-framework}

\subsection{Why Naive Evaluation Fails}
\label{sec:eval-naive}

Three evaluation approaches commonly applied to LLM diagnostic output
are inadequate for PROBE's multi-field structured outputs.

\paragraph{Cross-encoder semantic similarity truncates diagnostic
content.}
A natural evaluation pipeline relies on human-produced golden analysis, and uses a cross-encoder
model (\emph{e.g.,} RoBERTa-based, \texttt{stsb-roberta-large}) to score semantic
similarity between the text produced by the expert LLM and the golden text.  However, this model has a
512-token input window shared between both texts.  A typical
diagnostic output, containing frame-by-frame explanations, an issue
summary, and a chain-of-thought reasoning trace, commonly exceeds 800 tokens.
When two such outputs are compared, the tokenizer silently truncates
from the right, and the similarity score is computed on whatever fits
the window.

In practice, this means the score reflects whether two models
described the \emph{early} frames of a protocol exchange similarly,
while the diagnostic conclusion, root cause analysis, and late-frame
evidence (the content that matters most and that is often toward the end of the analysis text) are discarded.  Two models
that agree on the initial probe/authentication sequence but disagree
on whether the four-way handshake failed will score high on similarity
because the truncated window never reaches the handshake frames.

\paragraph{Similarity cannot distinguish agreement from
contradiction.}
Even within the token window, semantic similarity measures topical
relatedness, not factual consistency.  The sentences ``after reassociation, the four-way
handshake completed '' and ``after reassociation, the
four-way handshake was interrupted'' share
nearly identical vocabulary and sentence structure.  A cross-encoder may
assign them a high similarity score because they are \emph{about} the
same topic, despite reaching opposite diagnostic conclusions.
Fine-tuning on domain-specific paraphrase pairs (e.g., ``PSK'' 
$\leftrightarrow$ ``WPA2 password'') improves terminology matching
but does not address this fundamental limitation: similarity is
symmetric, but diagnostic agreement is directional.

\paragraph{Golden references co-produced by a model create circular
bias.}
The golden references in our evaluation corpus were produced by a
three-way process: a human SME provided annotations (that were considerated or discarded in the reconciliation phase, depending on their quality), Claude~Sonnet
generated an independent analysis, and Claude~Opus reconciled both
into a normalized golden output.  This means the golden reference
carries the analytical fingerprint of the models that produced it: 
how they group frames, what they emphasize, how they structure
reasoning.

When a different model (Model~B) is evaluated against this golden reference,
low scores may reflect stylistic divergence rather than diagnostic
error.  Model~B might group frames differently, emphasize different
protocol fields, or reach the same conclusion through a different
reasoning path (all of which reduce similarity scores without
indicating any loss of diagnostic quality).  Conversely, if the
\emph{same} model that produced the golden text is evaluated against that golden reference
(as in our Config~C), scores are inflated by the shared analytical
style.  Our ablation confirms this: Config~C (which mirrors the
golden generation pipeline) achieves Wt~$F_1 = 0.964$, while
Config~F (which uses a fundamentally different draft process) achieves
0.957, a gap that reflects evaluation circularity, not diagnostic
superiority (\Cref{sec:ablation}).

\subsection{Per-Field Assertion Matching}
\label{sec:eval-assertions}

Rather than comparing full-text outputs, we decompose each diagnostic
output into independently evaluable fields:

\begin{enumerate}[nosep,leftmargin=*]
\item \textbf{Frame coverage} (set comparison):
  $\text{Recall} = |\hat{\mathcal{F}} \cap \mathcal{F}^*| \;/\;
  |\mathcal{F}^*|$.
\item \textbf{Frame type agreement} (exact match on constrained
  literals): fraction of overlapping frames with matching protocol
  type.
\item \textbf{Explanation consistency} (LLM-judge with structured
  rubric, \emph{not} similarity): binary agreement on protocol
  behavior and outcome per matched frame group.
\item \textbf{Diagnostic conclusion} (LLM-judge):  ``Do these two
  summaries identify the same failure mode and the same cause?''
\end{enumerate}

\subsection{Weighted F-Beta with Tiered Importance}
\label{sec:eval-fbeta}

Golden evidence fields are assigned to diagnostic tiers:
\begin{itemize}[nosep,leftmargin=*]
\item \textbf{Tier~1} (weight $w_1$): fields that directly support
  the root-cause conclusion (e.g., EAPOL message types,
  deauthentication reason code, RSSI during handshake).
\item \textbf{Tier~2} (weight $w_2$): contextual evidence (e.g.,
  probe/association success, QoS indicators).
\item \textbf{Tier~3} (weight $w_3$): supplementary detail (e.g.,
  retransmission counts, timing).
\end{itemize}

Let $\hat{\mathcal{F}}_t$ be the set of predicted fields in tier~$t$
and $\mathcal{F}^*_t$ the corresponding golden reference set.  Let
$w_{\text{fp}}$ be the weight assigned to false positives (predicted
fields not in any golden tier).  The weighted true positives, false
negatives, and false positives are:

\begin{equation}
  \text{TP}_w = \sum_{t \in \{1,2,3\}} w_t \cdot
    |\hat{\mathcal{F}}_t \cap \mathcal{F}^*_t|
\label{eq:tp-weighted-eval}
\end{equation}

\begin{equation}
  \text{FN}_w = \sum_{t \in \{1,2,3\}} w_t \cdot
    |\mathcal{F}^*_t \setminus \hat{\mathcal{F}}_t|
\label{eq:fn-weighted-eval}
\end{equation}

\begin{equation}
  \text{FP}_w = w_{\text{fp}} \cdot
    \left|\hat{\mathcal{F}} \setminus
    \bigcup_{t} \mathcal{F}^*_t\right|
\label{eq:fp-weighted-eval}
\end{equation}

where $\hat{\mathcal{F}} = \hat{\mathcal{F}}_1 \cup
\hat{\mathcal{F}}_2 \cup \hat{\mathcal{F}}_3$ is the full set of
predicted fields.  The weighted precision and recall follow:

\begin{equation}
  P_w = \frac{\text{TP}_w}{\text{TP}_w + \text{FP}_w}
  \qquad
  R_w = \frac{\text{TP}_w}{\text{TP}_w + \text{FN}_w}
\label{eq:weighted-pr}
\end{equation}

with the convention that $P_w = 1$ when $\text{TP}_w + \text{FP}_w =
0$ (model predicts nothing) and $R_w = 1$ when $\text{TP}_w +
\text{FN}_w = 0$ (golden set is empty).  The weighted $F_\beta$ score
is:

\begin{equation}
  F_\beta = \frac{(1 + \beta^2) \cdot P_w \cdot R_w}
                 {\beta^2 \cdot P_w + R_w}
\label{eq:fbeta}
\end{equation}

The $\beta$ parameter controls the precision--recall tradeoff.  At
$\beta = 1$ (used in our experiments), precision and recall are
weighted equally.  Values $\beta > 1$ favor recall (penalizing missed
evidence more than spurious predictions), while $\beta < 1$ favor
precision (penalizing hallucinated evidence more than missed fields).
For diagnostic systems where a missed critical frame is more costly
than an extra contextual frame, $\beta = 1.5$ may be appropriate;
we use $\beta = 1$ as a conservative baseline.

\paragraph{Non-key weighting cap.}
In the specific case of frame-level evaluation, tier~2 frames
(relevant but non-key) can vastly outnumber tier~1 frames (key
diagnostic frames).  To prevent tier~2 from dominating the score
through sheer count, the effective weight for non-key frames is
capped:

\begin{equation}
  w_{\text{rel}} = \min\!\left(w_2,\;
    \frac{c \cdot |\mathcal{F}^*_1|}
         {|\mathcal{F}^*_2|}\right)
\label{eq:nonkey-cap}
\end{equation}

where $c$ is a cap multiplier (default $c = 1$) that limits the total
non-key contribution to at most $c \times |\mathcal{F}^*_1|$ weighted
units.  When the golden set has 2~key frames and 10~non-key relevant
frames, the uncapped tier~2 contribution would be $10 \times w_2$;
with $c = 1$, it is reduced to $2 \times w_{\text{rel}}$, ensuring
that key frame recovery remains the dominant signal.

\paragraph{Instantiation for frame highlighting.}
In our experiments (\Cref{sec:ablation}), we instantiate the general
framework for frame-level evaluation with $w_1 = 5$ (key frames),
$w_2 = 1$ (non-key relevant frames, subject to the cap), $w_3 = 0$
(no third tier), $w_{\text{fp}} = 1$, $c = 1$, and $\beta = 1$.
This produces the Wt~$F_1$ metric reported throughout the
experimental results.  The 5:1 ratio between key and non-key weights
reflects the diagnostic principle that missing a deauthentication
reason code (tier~1) is five times more consequential than missing a
routine probe response (tier~2).

\subsection{Assertion-Based Golden References}
\label{sec:eval-golden}

The golden references in our evaluation corpus, produced by a
three-way process described in \Cref{sec:eval-naive}, with an SME provided annotations, a Claude~Sonnet-generated independent draft, and Claude~Opus reconciliation into a normalized output
containing \texttt{suggestedGold} (summary, chain of thought, key
frames, relevant frames) and \texttt{llmReview} (structured
comparison of SME claims versus PCAP evidence).

This co-production process means the golden reference conflates two
distinct kinds of information:

\begin{enumerate}[nosep,leftmargin=*]
\item \textbf{PCAP-verifiable facts} that any correct analysis must
  recover, regardless of which model produced it.  These include:
  which frames contain the failure evidence (e.g., frame~126 contains
  an association response with status 0x0011), what protocol type
  those frames represent (e.g., EAPOL, Deauthentication), and what
  diagnostic conclusion the evidence supports (e.g., ``AP rejected
  association due to maximum client capacity'').
\item \textbf{Model-contributed narrative} that reflects how the
  co-producing models structured their reasoning.  This includes:
  how frames are grouped into sequences, which contextual frames are
  mentioned, the phrasing and level of detail in explanations, and
  the narrative arc of the chain of thought.
\end{enumerate}

Evaluating a new model against the full golden text penalizes
divergence in category~(2) as if it were an error in category~(1).
A model that correctly identifies frame~126 as the failure point but
describes the preceding authentication exchange differently from the
golden will score lower on text similarity, even though its
diagnosis is equally correct.

\paragraph{Extracting assertions.}
To separate verifiable facts from narrative style, we decompose each
golden reference into a set of \emph{assertions}: minimal, binary
claims that can be checked against the PCAP text without reference to
how any particular model would phrase them.  For each golden, the
assertion set includes:

\begin{itemize}[nosep,leftmargin=*]
\item \textbf{Required key frames}: the set of frame numbers in
  \texttt{suggestedGold.frames.key}.  Any correct analysis must
  identify these frames as diagnostically critical.  In our example
  (capture \texttt{fa:58:45}), the assertion is:
  $\mathcal{K}^* = \{120, 126\}$.
\item \textbf{Required relevant frames}: the set of frame numbers
  in \texttt{suggestedGold.frames.relevant}.  A correct analysis
  should cover these frames as protocol context, though missing one
  is less consequential than missing a key frame.
\item \textbf{Protocol type per frame group}: for each key frame,
  the protocol type is deterministically verifiable from the PCAP
  (e.g., frame~120 is a Disassociation, frame~126 is an Association
  Response).  These are exact-match assertions requiring no NLP.
\item \textbf{Diagnostic conclusion}: a short factual statement of
  the root cause, derived from the SME's annotation and verified by
  the reconciler.  For evaluation, this is checked via an LLM judge
  asking: ``Does this analysis identify the same failure mode and
  the same cause as the reference?'' (a binary consistency check,
  not a similarity score).
\end{itemize}

\paragraph{What assertions exclude.}
The assertion set deliberately excludes: the narrative structure (how the
chain of thought is organized), the phrasing (whether the model says
``handshake timeout'' or ``EAPOL exchange did not complete''),
the contextual elaboration (how much detail is provided about normal
frames preceding the failure), and any recommendation content (what
remediation steps are suggested).  These are legitimate dimensions of
output quality but they are \emph{style}, not diagnosis \emph{correctness}, and
scoring them penalizes models that reason differently from the golden
co-producer without any diagnostic benefit.

\paragraph{Practical construction.}
Building assertion sets does not require re-annotating the dataset
from scratch.  The key and relevant frame sets are already present in
every golden file (\texttt{suggestedGold.frames}).  Protocol types are
deterministic given the frame number and the PCAP.  The diagnostic
conclusion can be extracted from the existing
\texttt{suggestedGold.pcapSummary} by an SME confirming the factual
core in one sentence.  For the 104~cases in our evaluation corpus,
the frame-level assertions are used directly; diagnostic conclusion
assertions are validated through the reconciler's structured review
(\texttt{llmReview.therefore}), which explicitly states the logical
connection between evidence and conclusion.

\paragraph{Evaluation against assertions.}
With assertion-based references, the evaluation becomes:
\begin{enumerate}[nosep,leftmargin=*]
\item \textbf{Frame coverage} (set comparison): Does the model's
  output cover the required key and relevant frames?  Scored via
  weighted $F_\beta$ (\Cref{sec:eval-fbeta}).
\item \textbf{Protocol type agreement} (exact match): For covered
  frames, does the model assign the correct protocol type?  Binary
  per frame, reported as agreement rate.
\item \textbf{Diagnostic consistency} (LLM judge, binary): Does the
  model's conclusion match the assertion's root cause?  Checked by
  asking: ``Do these identify the same failure mode?''
\end{enumerate}

This three-level evaluation is model-agnostic: it scores any model's
output against PCAP-verifiable facts, not against how a specific
co-producing model happened to phrase its analysis.  In our
experiments, the frame coverage component (levels~1 and~2) is used
throughout; diagnostic consistency (level~3) is deferred to future
work pending construction of per-case root-cause assertion labels.

\section{Experimental Results}
\label{sec:experiments}

We evaluate the PROBE pipeline through five complementary experiments:
a progressive ablation of pipeline components (\Cref{sec:ablation}),
an analysis of ensemble dimensionality (\Cref{sec:ensemble-dim}),
a confidence calibration study (\Cref{sec:calibration}),
a verdict accuracy analysis (\Cref{sec:verdict}), and
an efficiency analysis (\Cref{sec:efficiency}).
All experiments use the same corpus of 104~capture--reviewer pairs
spanning 87~unique 802.11 captures btained from real networks where users reported potential issues, and spanning across multiple protocol failure categories (EAPOL
handshake failures, association rejections, DHCP failures, roaming
anomalies, and deauthentication events). While 70 captures were reviewed by a single human SME, 17 captures were selected for having been reviewed by 2, independent, human SMEs, and were treated as independent capture-reviewer pairs.

\subsection{Pipeline Ablation}
\label{sec:ablation}

To quantify the contribution of each pipeline stage, we evaluate six
configurations against the capture dataset, progressively
adding components from the simplest baseline (human SME annotations
alone) to the full PROBE pipeline.  All configurations are evaluated
on the same corpus of 104~capture--reviewer pairs.

\subsubsection{Configurations}
\label{sec:ablation-configs}

We define six configurations organized into two tiers: \emph{generation-only}
(no reconciliation step) and \emph{with reconciliation} (an additional judge
model synthesizes the final output):

\paragraph{Tier~1: Generation only.}
\begin{itemize}[nosep,leftmargin=*]
\item \textbf{Config~A (SME only).}  A human subject-matter expert has generated annotations that
  are used as the diagnosis.  No LLM is involved.  SME
  annotations typically contain a short textual summary, a chain of
  thought, and often one or more highlighted frame numbers or fields. These annotations, extracted from real-field documents, serve as the human expert baseline.
\item \textbf{Config~B (Sonnet one-shot).}  A single Claude~Sonnet~4.5
  call with one run and one candidate.  This configuration is the common choice when subcontracting network troubleshooting tasks to an LLM. The model receives the
  full PCAP text representation and produces a structured diagnosis
  including relevant frames, evidence items with
  contributing/non-contributing annotations, and a verdict from a
  strict taxonomy (\texttt{CONFIRMED\_ISSUE}, \texttt{PLAUSIBLE\_ISSUE},
  \texttt{INSUFFICIENT\_EVIDENCE}, \texttt{NO\_ISSUE\_FOUND}).
  This configuration uses no reconciliation, no ensemble diversity, no SME input.
\item \textbf{Config~D (Ensemble $3\!\times\!3$).}  Three independent
  Sonnet runs, each producing three candidate diagnoses, for a total of
  nine candidates per capture.  The selected output is determined by
  majority voting on verdict and diagnosis, with the best-scoring
  candidate from the majority group chosen as the final prediction.
  No reconciliation, no second opinion model.
\item \textbf{Config~E (Ensemble + second opinion).}  Same $3\!\times\!3$
  Sonnet ensemble as Config~D, plus an independent analysis from
  Llama~3.3~70B via Bedrock inference profile.  The second opinion
  model has no visibility into the Sonnet ensemble.  Candidate selection
  still uses majority voting.  No reconciliation.
\end{itemize}

\paragraph{Tier~2: With reconciliation.}
\begin{itemize}[nosep,leftmargin=*]
\item \textbf{Config~C (One-shot + reconcile).}  A single Sonnet draft
  followed by Claude~Opus~4.1 reconciliation.  The reconciler receives
  the PCAP text, the SME annotation, and the single draft output, then
  produces a normalized golden diagnosis with a structured comparison
  of SME claims versus PCAP evidence, and model claims versus PCAP
  evidence.  This configuration closely mirrors the pipeline that
  originally produced the golden references.
\item \textbf{Config~F (Full PROBE pipeline).}  The complete system:
  $3\!\times\!3$ Sonnet ensemble with Llama~3.3~70B second opinion,
  followed by Opus reconciliation.  The reconciler sees all nine
  ensemble candidates, the second opinion output, and the SME annotation,
  and synthesizes the final diagnosis.
\end{itemize}

\subsubsection{Evaluation Metrics}
\label{sec:ablation-metrics}

Each configuration output is scored against the golden reference on
frame-level evidence quality using four complementary metrics:

\paragraph{Key frame precision and recall.}  Key frames are the
diagnostically critical frames that directly provide supporting evidence for the root cause diagnosis
(e.g., a deauthentication frame with a specific reason code, an EAPOL
message showing handshake failure).  Key frame precision measures what
fraction of the predicted key frames match the golden key set; recall
measures what fraction of the golden key frames are recovered.

\paragraph{Relevant frame precision and recall.}  Relevant frames
include the full protocol exchange context: probes, authentication,
association, and handshake messages (802.11 and/or application-related).  These elements provide the narrative context,
but are less diagnostically decisive than key frames. They are important to install the narrative context within which an issue was observed (e.g., first association or reassociation while roaming).

\paragraph{Weighted $F_1$ (Wt\_F1).}  We compute a
weighted $F_1$ that assigns weight $w_{\text{key}} = 5$ to key frames
and weight $w_{\text{rel}} = 1$ to non-key relevant frames, with
false positives penalized at weight $w_{\text{fp}} = 1$.  This
ensures that missing a diagnostically critical frame is penalized
five times more heavily than missing a contextual frame:
\begin{equation}
  \text{TP}_w = w_{\text{key}} \cdot |P \cap K| + w_{\text{rel}} \cdot |P \cap (R \setminus K)|
\label{eq:tp-weighted}
\end{equation}
\begin{equation}
  \text{FN}_w = w_{\text{key}} \cdot |K \setminus P| + w_{\text{rel}} \cdot |(R \setminus K) \setminus P|
\label{eq:fn-weighted}
\end{equation}
\begin{equation}
  \text{FP}_w = w_{\text{fp}} \cdot |P \setminus R|
\label{eq:fp-weighted}
\end{equation}
where $P$ is the set of predicted frames, $K$ the golden key frames,
and $R$ the golden relevant frames ($K \subseteq R$).  Weighted
precision and recall follow, with $F_1$ as their harmonic mean.

\paragraph{Perfect key match rate.}  The fraction of cases where the
predicted key frame set exactly matches the golden key frame set
($F_1 = 1.0$).  This stringent metric captures whether the system
identifies precisely the right diagnostic evidence without omission or
false inclusion.

\subsubsection{Results}

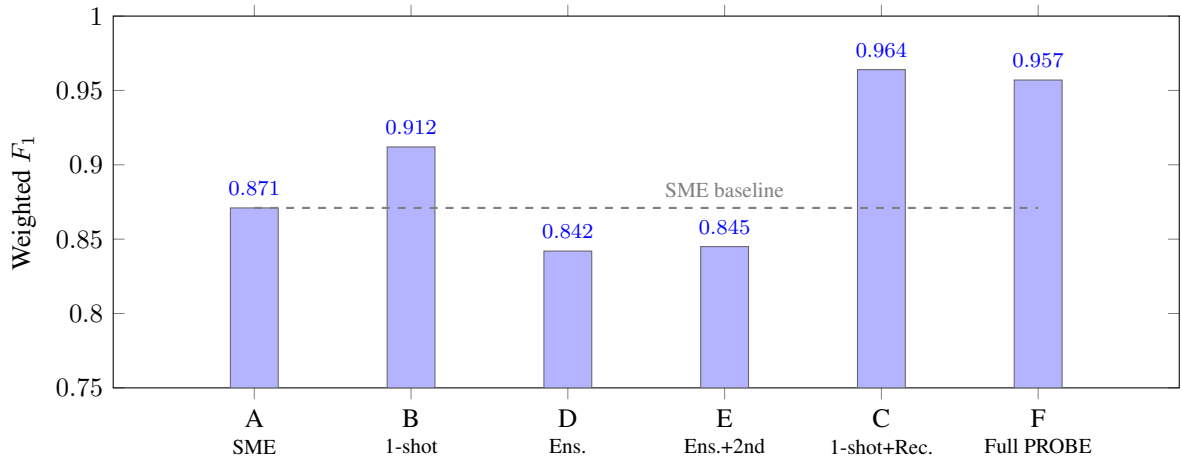
\begin{figure}[t]
\centering
\begin{tikzpicture}
\begin{axis}[
    ybar,
    width=0.95\linewidth,
    height=6.5cm,
    bar width=18pt,
    ylabel={Weighted $F_1$},
    ylabel style={font=\small},
    symbolic x coords={A,B,D,E,C,F},
    xtick=data,
    xticklabels={%
      {A\\[-2pt]\scriptsize SME},
      {B\\[-2pt]\scriptsize 1-shot},
      {D\\[-2pt]\scriptsize Ens.},
      {E\\[-2pt]\scriptsize Ens.+2nd},
      {C\\[-2pt]\scriptsize 1-shot+Rec.},
      {F\\[-2pt]\scriptsize Full PROBE}},
    xticklabel style={align=center,font=\small},
    ymin=0.75, ymax=1.0,
    ytick={0.75,0.80,0.85,0.90,0.95,1.00},
    yticklabel style={font=\small},
    nodes near coords,
    nodes near coords style={
        font=\scriptsize,
        yshift=1pt,
        /pgf/number format/.cd, fixed, fixed zerofill, precision=3
    },
    enlarge x limits=0.18,
    visualization depends on={value \thisrow{barcolor} \as \barcolor},
]

\addplot+[
    ybar,
    bar shift=0pt,
    draw=black!60,
    point meta=explicit symbolic,
    nodes near coords={\pgfmathprintnumber[fixed,fixed zerofill,precision=3]\pgfplotspointmeta},
] table [x=cfg, y=val, meta=val, col sep=comma, row sep=\\] {
cfg, val, barcolor \\
A, 0.871, 1 \\
B, 0.912, 2 \\
D, 0.842, 3 \\
E, 0.845, 4 \\
C, 0.964, 5 \\
F, 0.957, 6 \\
};

\draw[dashed, gray, thick]
  (axis cs:A,0.871) -- (axis cs:F,0.871);

\node[gray, font=\scriptsize, anchor=south]
  at (axis cs:E,0.871) {SME baseline};

\end{axis}
\end{tikzpicture}
\caption{Weighted $F_1$ by configuration. Generation-only (A, B, D, E)
  vs.\ with reconciliation (C, F). Dashed line marks the SME baseline.
  Ensemble without reconciliation (D, E) falls \emph{below} the SME
  baseline. Both reconciled configurations (C, F) substantially exceed
  all generation-only results.}
\label{fig:ablation-bar}
\end{figure}

\Cref{tab:ablation} presents the complete results.
\Cref{fig:ablation-bar} visualizes the primary metric (Wt~$F_1$)
across configurations.

\begin{table}[t]
  \centering
  \caption{Pipeline ablation.  \textbf{Wt~$F_1$} is the primary metric
    (key frames weighted $5\times$).  Configs~C/F have $N\!=\!100$
    due to 4~reconciler JSON failures.}
  \label{tab:ablation}
  \small
  \begin{tabular}{@{}clcccccccc@{}}
    \toprule
    & \textbf{Config}
      & \textbf{Desc.}
      & \textbf{$N$}
      & \textbf{Key P}
      & \textbf{Key R}
      & \textbf{Key $F_1$}
      & \textbf{Rel $F_1$}
      & \textbf{Wt~$F_1$}
      & \textbf{Perf.~Key} \\
    \midrule
    \multirow{4}{*}{\rotatebox{90}{\scriptsize Gen.}}
    & A & SME only                 & 104 & .747 & \textbf{.915} & .773 & .460 & .871 & 45.2\% \\
    & B & Sonnet 1-shot            & 104 & .860 & .665          & .680 & .831 & .912 & 34.6\% \\
    & D & Ensemble $3\!\times\!3$  & 104 & .711 & .490          & .444 & .692 & .842 & 21.2\% \\
    & E & Ens.\ + 2nd opinion      & 104 & .720 & .502          & .461 & .692 & .845 & 20.2\% \\
    \arrayrulecolor{tierline}\midrule\arrayrulecolor{black}
    \multirow{2}{*}{\rotatebox{90}{\scriptsize +Rec.}}
    & C & 1-shot + reconcile       & 100 & \textbf{.941} & .786 & \textbf{.815} & \textbf{.851} & \textbf{.964} & \textbf{51.0\%} \\
    & F & Full PROBE               & 100 & .932 & .757          & .776 & .846 & .957 & 50.0\% \\
    \bottomrule
  \end{tabular}
\end{table}

\subsubsection{Key Findings}

\paragraph{SME experts excel at diagnostic frames but lack coverage.}
Config~A achieves the highest key frame recall (0.915): experts jump
to the diagnostic evidence when it is obvious (deauthentication reason codes, rejected
associations with status codes) but omit the surrounding protocol context
(Rel~Rec~=~0.342).

\paragraph{Single-pass LLM analysis inverts this pattern.}
Config~B raises relevant recall from 0.342 to 0.818 but drops key
recall from 0.915 to 0.665.  The model narrates the full exchange with ease, but often makes conservative conclusions, and
misses the diagnostic punchline in one-third of cases.

\paragraph{Ensemble without reconciliation is counterproductive.}
Config~D (Wt~$F_1$~=~0.842) falls \emph{below} the SME baseline
(0.871).  Majority voting amplifies conservative verdicts: three cases
score 0.000 where nine candidates unanimously converge on
\texttt{NO\_ISSUE\_FOUND} for captures with confirmed failures.
Adding a second opinion model (Config~E, 0.845) provides no
meaningful improvement ($+$0.003). This finding is central, because without reconciliation, a natural conclusion would be that multiple opinions, from one or more models, do not bring value. However, reconciliation will show how much multiple opinions are needed and why relying on a single opinion, as shown above, fails.

\paragraph{Reconciliation is the decisive component, but benefits
from ensemble diversity.}
The reconcile step transforms performance: Config~C reaches 0.964,
Config~F reaches 0.957.  However, the 0.7\% gap between C and F
\emph{overstates} Config~C's advantage because Config~C closely
mirrors the pipeline that produced the golden references (a single
Sonnet draft reconciled by Opus).  Config~F achieves near-identical
quality through a fundamentally different reasoning path, one that
is not anchored to the golden's generation process.

More importantly, the full pipeline provides three capabilities
that Config~C cannot:
\begin{enumerate}[nosep,leftmargin=*]
\item \textbf{Robustness on hard cases.}  When the single draft in
  Config~C happens to miss the diagnostic frames, the reconciler has
  no alternative hypothesis to recover from.  Config~F's reconciler
  sees nine candidates plus a second opinion, enabling recovery of
  correct minority diagnoses that the single draft missed.
  Config~F's second-worst case (Wt~$F_1$~=~0.706) identifies all key
  frames correctly; Config~F's single outlier (0.231) involves a
  capture that also challenges Config~C.
\item \textbf{Measurable diagnostic diversity.}  The ensemble
  generates stability and agreement signals that enable reliability
  assessment (\Cref{sec:calibration}).  Config~C produces a single
  draft and a single reconciliation with no internal signal of how
  confident the system should be.
\item \textbf{Independence from the golden generation process.}
  As the evaluation corpus grows beyond captures whose golden
  references were produced by Config~C's pipeline, the circularity
  advantage disappears.  Config~F's architecture is designed to
  generalize; Config~C's advantage is structural to this specific
  evaluation.
\end{enumerate}

\subsubsection{Worst-Case Behavior}

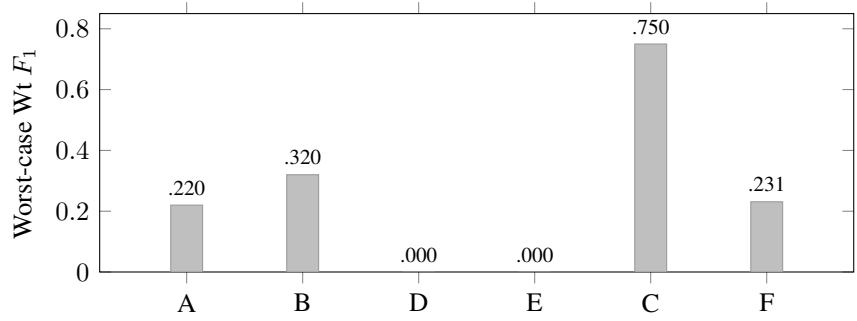
\begin{figure}[t]
\centering
\begin{tikzpicture}
\begin{axis}[
    ybar,
    width=0.7\linewidth,
    height=5cm,
    bar width=12pt,
    ylabel={Worst-case Wt $F_1$},
    ylabel style={font=\small},
    symbolic x coords={A,B,D,E,C,F},
    xtick=data,
    xticklabels={A,B,D,E,C,F},
    xticklabel style={font=\small},
    ymin=0, ymax=0.85,
    yticklabel style={font=\small},
    nodes near coords,
    every node near coord/.append style={font=\scriptsize,above},
    point meta=explicit symbolic,
    enlarge x limits=0.15,
]
\addplot[fill=gray!50, draw=gray!80]
  coordinates {
    (A,0.220) [{.220}]
    (B,0.320) [{.320}]
    (D,0.000) [{.000}]
    (E,0.000) [{.000}]
    (C,0.750) [{.750}]
    (F,0.231) [{.231}]
  };
\end{axis}
\end{tikzpicture}
\caption{Worst-case Wt~$F_1$ per configuration.  Ensemble without
  reconciliation (D,~E) produces catastrophic failures
  (Wt~$F_1\!=\!0$).  Config~C has the highest floor (0.750).
  Config~F has one outlier (0.231) but its second-worst case is 0.706.}
\label{fig:worst-case}
\end{figure}

\Cref{fig:worst-case} compares the worst-case Wt~$F_1$ (floor) and
the count of catastrophic failures (Wt~$F_1 = 0.000$) across
configurations.

Generation-only ensembles (D,~E) produce three catastrophic misses
each (Wt~$F_1 = 0.000$).  Config~C eliminates all catastrophic
failures entirely (floor~=~0.750).  Config~F has one outlier at 0.231
(capture 2E:91, which also challenges Configs~A, B, and E), but its
second-worst case is 0.706 with perfect key frame identification.

\subsection{Ensemble Dimensionality}
\label{sec:ensemble-dim}

The ensemble operates along two orthogonal diversity axes: \emph{run
diversity} ($N$~independent calls with prompt variants, testing
reasoning stability) and \emph{candidate diversity} ($M$~hypotheses
per call, testing internal ambiguity).  Total candidates are
$N \times M$.

The ablation provides two key observations about dimensionality:

\paragraph{Run diversity matters more than candidate diversity for
reliability.}  Stability metrics are computed across runs, not within
them.  Candidates from the same run share prompt context and are not
statistically independent.  For reliability scoring, $(3\!\times\!1)$
is more informative than $(1\!\times\!3)$.

\paragraph{Diminishing returns under reconciliation.}  Comparing
Config~B ($1\!\times\!1$) to Config~F ($3\!\times\!3$), both with
reconciliation: Wt~$F_1$ moves from 0.964 to 0.957, effectively
flat.  The reconciler is already the binding quality constraint.
Additional ensemble diversity adds reliability signals and robustness
on edge cases, but does not meaningfully improve the reconciler's
already-high selection quality.

\Cref{tab:cost} summarizes the cost--quality tradeoff.

\begin{table}[h]
  \centering
  \caption{API calls and quality by configuration.  Config~C is
    most cost-efficient; Config~F adds reliability measurement at
    2.5$\times$ the call cost.}
  \label{tab:cost}
  \small
  \begin{tabular}{@{}llcccc@{}}
    \toprule
    \textbf{Config} & \textbf{Description}
      & \textbf{Calls} & \textbf{Wt~$F_1$} & \textbf{Reliability?} \\
    \midrule
    B & One-shot           & 1 & .912 & No \\
    D & Ensemble           & 3 & .842 & Partial \\
    C & 1-shot + reconcile & 2 & .964 & No \\
    F & Full PROBE         & 5 & .957 & Yes \\
    \bottomrule
  \end{tabular}
\end{table}

The choice between Config~C and Config~F depends on whether
downstream processes require a reliability signal.  When only the
best-quality diagnosis is needed, Config~C dominates on cost
efficiency.  When automated quality assessment, human-review routing,
or capture difficulty characterization are required, Config~F provides
the necessary ensemble-derived signals at a 2.5$\times$ cost premium.

\subsection{Confidence Calibration}
\label{sec:calibration}

A central claim of PROBE is that diagnostic reliability can be
assessed through deterministic signals rather than LLM self-report.
We evaluate two confidence measures against actual correctness.

\subsubsection{Self-Reported Confidence Is Uninformative}

\begin{figure}[t]
\centering
\begin{tikzpicture}
\begin{axis}[
    ybar,
    width=0.75\linewidth,
    height=4.5cm,
    bar width=10pt,
    ylabel={Count},
    ylabel style={font=\small},
    xlabel={Self-reported confidence},
    xlabel style={font=\small},
    xtick={0.65,0.75,0.85,0.90,0.92,0.95,1.00},
    xticklabel style={font=\scriptsize},
    yticklabel style={font=\small},
    ymin=0,
    enlarge x limits=0.08,
    legend style={at={(0.98,0.98)},anchor=north east,font=\scriptsize},
]
\addplot[fill=sonettorange!60, draw=sonettorange!80]
  coordinates {(0.65,1) (0.75,2) (0.85,10) (0.90,14) (0.92,3) (0.95,74) (1.00,0)};
\addplot[fill=reconcpurple!40, draw=reconcpurple!70]
  coordinates {(0.65,0) (0.75,0) (0.85,4) (0.90,1) (0.92,0) (0.95,89) (1.00,9)};
\legend{Draft (Sonnet), Gold (Opus)}
\end{axis}
\end{tikzpicture}
\caption{Self-reported confidence distribution.  71\% of draft values
  and 86\% of gold values are exactly 0.95.  The distribution is
  effectively a point mass, providing no discriminative signal.}
\label{fig:self-conf-hist}
\end{figure}
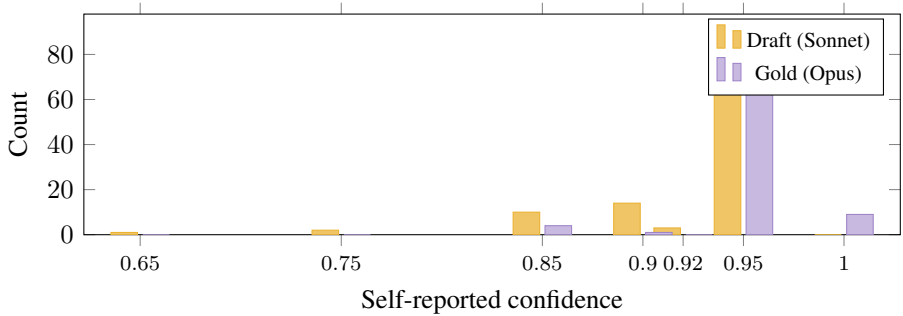

\Cref{fig:self-conf-hist} shows the distribution of self-reported
confidence from the 104~golden review files.

With 71\% of draft values at exactly 0.95 regardless of diagnostic
difficulty, self-reported confidence cannot distinguish routine
captures from genuinely ambiguous ones.  Spearman correlation with
correctness is $\rho = +0.35$ (Config~B), AUROC for detecting cases
requiring review is 0.57, barely above random.

\subsubsection{Composite Confidence: Architecture vs.\ Calibration}

\begin{figure}[t]
\centering
\begin{tikzpicture}
\begin{axis}[
    ybar interval,
    width=0.75\linewidth,
    height=4.5cm,
    ylabel={Count},
    ylabel style={font=\small},
    xlabel={Composite confidence},
    xlabel style={font=\small},
    xticklabel style={font=\scriptsize},
    yticklabel style={font=\small},
    ymin=0,
    xtick={0.0,0.1,0.2,0.3,0.4,0.5,0.6,0.7},
    xmin=0, xmax=0.7,
]
\addplot[fill=fullpipe!50, draw=fullpipe!80]
  coordinates {(0.0,24) (0.1,0) (0.2,33) (0.3,26) (0.4,9) (0.5,10) (0.6,1) (0.7,0)};
\end{axis}
\end{tikzpicture}
\caption{Composite confidence distribution (103~cases).  Range
  $[0.075, 0.625]$, mean~0.293.  Unlike self-reported confidence, the
  composite spreads across the full range, but as we show below,
  spread alone does not guarantee calibration.}
\label{fig:composite-hist}
\end{figure}
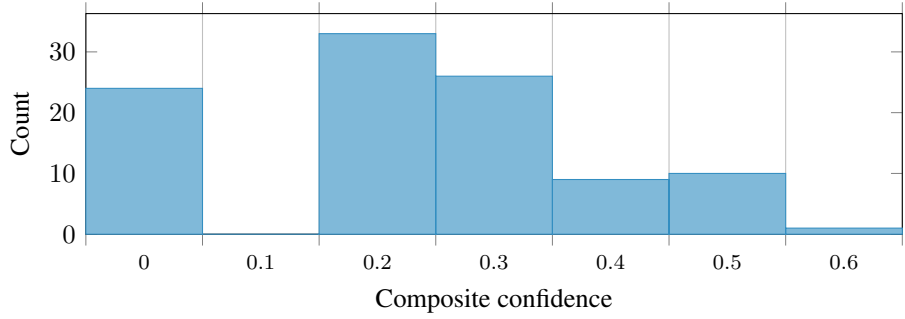

The deterministic composite is computed from evidence validity,
run-to-run stability, and cross-model agreement across the
$3\!\times\!3$ ensemble.  \Cref{fig:composite-hist} shows its
distribution.

Unlike self-reported confidence, the composite spans a wide range
(0.075--0.625).  However, calibration analysis reveals it is
\emph{not} predictive of correctness in the current evaluation
(\Cref{tab:cal-comparison}).

\begin{table}[t]
  \centering
  \caption{Confidence predictiveness.  Neither measure reliably
    predicts correctness, but for structurally different reasons.}
  \label{tab:cal-comparison}
  \small
  \begin{tabular}{@{}llccc@{}}
    \toprule
    \textbf{Confidence} & \textbf{Predicting}
      & \textbf{$\rho$} & \textbf{AUROC} & \textbf{Err.\ rate} \\
    \midrule
    Self-reported  & Config B & $+.35$ & 0.57 & 16.3\% \\
    Composite      & Config D & $+.07$ & 0.55 & 25.0\% \\
    Composite      & Config F & $+.09$ & 0.41 & 4.0\% \\
    \bottomrule
  \end{tabular}
\end{table}

\subsubsection{Why Calibration Fails, and Why That Is Good News}

The composite's poor calibration stems from two distinct mechanisms
depending on which pipeline stage it is evaluated against.

\paragraph{Against Config~D (raw ensemble, 25\% error rate):}
High ensemble agreement often indicates unanimous convergence on the same verdict, even when that is the 
\emph{wrong} verdict.  The ensemble confidently agrees on
\texttt{NO\_ISSUE\_FOUND} for captures with confirmed failures,
inverting the expected confidence--correctness relationship. This is because models tend to prefer flat description of the exchanges with conservative conclusions.

\begin{figure}[t]
\centering

\begin{tikzpicture}
\begin{axis}[
    width=0.75\textwidth,
    height=6cm,
    xlabel={Mean confidence bucket},
    ylabel={Accuracy, Wt F1 >= 0.8},
    xlabel style={font=\small},
    ylabel style={font=\small},
    xmin=0, xmax=1.05,
    ymin=0, ymax=1.1,
    xtick={0,0.2,0.4,0.6,0.8,1.0},
    ytick={0,0.2,0.4,0.6,0.8,1.0},
    xticklabel style={font=\small},
    yticklabel style={font=\small},
    legend style={
        at={(0.02,0.02)},
        anchor=south west,
        font=\scriptsize
    },
    grid=major,
    grid style={gray!30}
]

\addplot[dashed, gray, thick, domain=0:1] {x};
\addlegendentry{Perfect calibration}

\addplot[
    mark=square*,
    mark size=3pt,
    orange,
    thick
]
coordinates {
    (0.847,0.500)
    (0.928,0.400)
    (0.950,0.800)
    (0.950,1.000)
    (0.950,1.000)
};
\addlegendentry{Self-reported}

\addplot[
    mark=*,
    mark size=3pt,
    blue,
    thick
]
coordinates {
    (0.125,1.000)
    (0.215,0.950)
    (0.336,1.000)
    (0.467,0.900)
    (0.596,0.950)
};
\addlegendentry{Composite to Config F}

\end{axis}
\end{tikzpicture}

\caption{Calibration plot. The dashed diagonal represents perfect
calibration. Self-reported confidence, shown as orange squares, clusters
near x = 0.95 with variable accuracy; it cannot distinguish easy from
hard cases. Composite confidence against Config~F, shown as blue circles,
spreads along the x-axis, but accuracy is uniformly at least 0.90. The
reconciler eliminates most errors regardless of ensemble agreement,
leaving no failures to predict.}
\label{fig:cal-plot}
\end{figure}
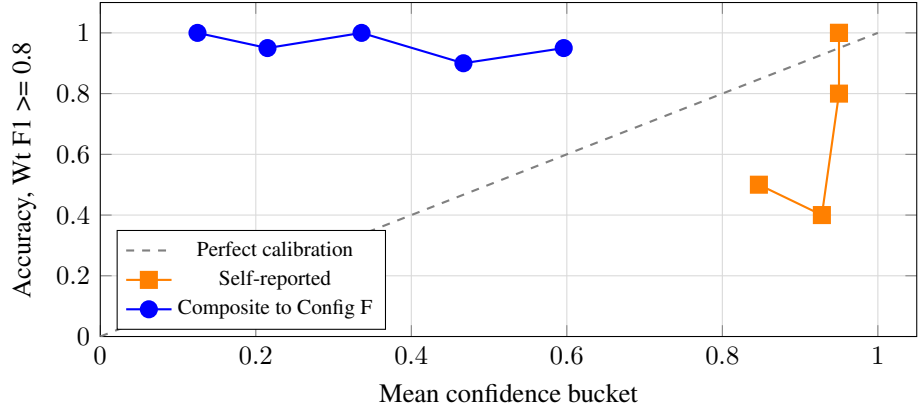

\paragraph{Against Config~F (full pipeline, 4\% error rate):}
The reconciler is so effective that correctness is uniformly high
across all confidence levels (\Cref{fig:cal-plot}). With only 4
errors in 100~cases, there is insufficient variance for any signal to
predict.

This finding is, paradoxically, a positive result for the PROBE
architecture.  It means the reconciler is robust enough to produce
correct diagnoses even when the ensemble is internally conflicted.
The 4\% error rate of Config~F, uniform across all confidence
levels, represents the current performance ceiling, not a
calibration failure.

\subsubsection{Reframing Confidence: Quality Gating vs.\ Capture Characterization}
\label{sec:cal-reframe}

These results reframe the role of confidence in the PROBE pipeline.
Rather than gating output quality (``is this answer correct?''),
composite confidence serves as a \emph{capture difficulty indicator}
(``is this capture inherently ambiguous?'').

Low ensemble agreement does not predict that the pipeline will produce
a wrong answer.  Instead, it identifies captures where the underlying
diagnostic question may be inherently underdetermined, for example those with truncated
exchanges, edge-case protocols, ambiguous timing, and exchanges of type 3, where the observed issue results from a failure that occurred outside of the capture point vantage point.  These captures
merit human review not for error correction (Config~F is correct 96\%
of the time) but for \emph{dataset enrichment}: they represent the
edge cases that could improve future models.

Component-level analysis supports this interpretation.  Of the 92
low-confidence cases (composite $< 0.5$), evidence validity is below
threshold in 100\%, stability in 65--72\%, and cross-model agreement
in 65\%.  Evidence validity is systematically depressed because the
current prompt does not consistently produce \texttt{contributes=true}
evidence items for non-failure verdicts, a prompt engineering
artifact that the verdict-aware evidence rules (\Cref{sec:ensemble})
are designed to address.  Fixing this component and expanding the
evaluation to more difficult captures (where the reconciler's 4\%
error rate will increase) are the two paths to achieving meaningful
calibration.

\paragraph{Practical deployment implications.}
The pipeline can \textbf{auto-accept} 96\% of cases without
confidence-based gating. Low confidence cases are easily identified and labeled. These low-confidence cases can be routed to
human review not because the pipeline is likely wrong, but because the
capture is likely \emph{interesting}.  Self-reported confidence should
not be used for any decision-making purpose.

\subsection{Verdict Accuracy and False Negative Analysis}
\label{sec:verdict}
\label{sec:truncation}

The PROBE pipeline introduces a strict four-value verdict taxonomy
(\texttt{CONFIRMED\_ISSUE}, \texttt{PLAUSIBLE\_ISSUE},
\texttt{INSUFFICIENT\_EVIDENCE}, \texttt{NO\_ISSUE\_FOUND}) to
prevent hallucinated diagnoses on ambiguous or truncated captures.
This section evaluates how accurately each generation-only
configuration assigns verdicts, and quantifies the rate at which
models fail to identify confirmed protocol failures.

\subsubsection{Dataset Composition}

All 104 golden references in our evaluation corpus contain confirmed
protocol failures (100\% \texttt{CONFIRMED\_ISSUE}).  This dataset
composition reflects the annotation workflow: SMEs were asked to
review captures exhibiting known connectivity problems (because they were extracted from real-life networks were support tickets were the cause of the PCAP routing to human review), not to label
captures that completed normally.

This composition has an important implication for the analysis: false
positive rate (the model hallucinating an issue where none exists)
\emph{cannot be measured} because the dataset contains no negative
cases.  The measurable error is exclusively \emph{false negatives}:
the model concluding \texttt{NO\_ISSUE\_FOUND} or
\texttt{INSUFFICIENT\_EVIDENCE} on a capture with a confirmed failure.
We return to this limitation in \Cref{sec:verdict-limitations}.

\subsubsection{Verdict Confusion Matrices}

\Cref{tab:verdict-confusion} presents the verdict confusion matrices
for the three generation-only configurations that produce explicit
verdicts (Configs~B, D, E).  Reconciled configurations (C, F) produce
a normalized golden without an explicit verdict from the draft
taxonomy and are therefore excluded from this analysis.

\begin{table}[t]
  \centering
  \caption{Verdict confusion matrices for generation-only
    configurations.  All 104 golden cases are
    \texttt{CONFIRMED\_ISSUE} (rows).  Columns show predicted
    verdicts.  Correct predictions are in bold.}
  \label{tab:verdict-confusion}
  \small
  \begin{tabular}{@{}llrrrr@{}}
    \toprule
    & & \multicolumn{4}{c}{\textbf{Predicted verdict}} \\
    \cmidrule(l){3-6}
    \textbf{Config} & \textbf{Golden}
      & \texttt{CONF.} & \texttt{PLAUS.}
      & \texttt{INSUF.} & \texttt{NO\_ISS.} \\
    \midrule
    B & \texttt{CONFIRMED} & \textbf{81} & 1  & 4  & 18 \\
    D & \texttt{CONFIRMED} & \textbf{30} & 22 & 26 & 26 \\
    E & \texttt{CONFIRMED} & \textbf{31} & 21 & 29 & 23 \\
    \bottomrule
  \end{tabular}
\end{table}

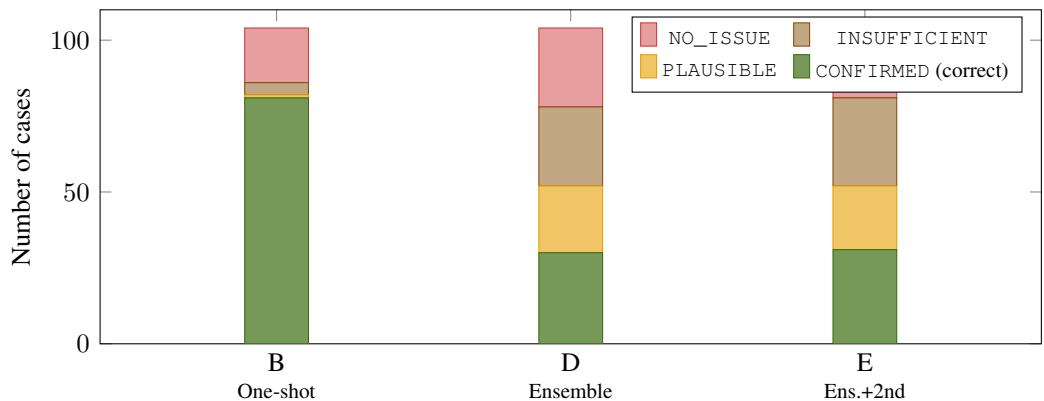
\begin{figure}[t]
\centering
\begin{tikzpicture}
\begin{axis}[
    ybar stacked,
    width=0.85\linewidth,
    height=6cm,
    bar width=24pt,
    ylabel={Number of cases},
    ylabel style={font=\small},
    symbolic x coords={B,D,E},
    xtick=data,
    xticklabels={%
      {B\\[-2pt]\scriptsize One-shot},
      {D\\[-2pt]\scriptsize Ensemble},
      {E\\[-2pt]\scriptsize Ens.+2nd}},
    xticklabel style={align=center, font=\small},
    ymin=0, ymax=110,
    yticklabel style={font=\small},
    enlarge x limits=0.3,
    legend style={
      at={(0.98,0.98)}, anchor=north east,
      font=\scriptsize,
      legend columns=2,
      /tikz/every even column/.append style={column sep=4pt},
    },
    reverse legend,
]
\addplot[fill=greenstroke!70, draw=greenstroke]
  coordinates {(B,81) (D,30) (E,31)};
\addplot[fill=sonettorange!60, draw=sonettorange]
  coordinates {(B,1) (D,22) (E,21)};
\addplot[fill=amberstroke!50, draw=amberstroke]
  coordinates {(B,4) (D,26) (E,29)};
\addplot[fill=ensred!50, draw=ensred]
  coordinates {(B,18) (D,26) (E,23)};
\legend{%
  \texttt{CONFIRMED} (correct),
  \texttt{PLAUSIBLE},
  \texttt{INSUFFICIENT},
  \texttt{NO\_ISSUE}%
}
\end{axis}
\end{tikzpicture}
\caption{Verdict distribution by configuration.  All 104 golden cases
  are confirmed issues.  Green segments represent correct verdict
  assignment.  Single-pass Sonnet (Config~B) correctly identifies
  78\% of confirmed issues; the ensemble (Configs~D,~E) drops to
  29--30\%.  The ensemble's majority-vote mechanism amplifies
  conservative verdicts (\texttt{INSUFFICIENT\_EVIDENCE} and
  \texttt{NO\_ISSUE\_FOUND}), which together account for 50\% of
  ensemble predictions.}
\label{fig:verdict-distribution}
\end{figure}

\Cref{fig:verdict-distribution} visualizes the verdict distribution
across configurations.

\subsubsection{False Negative Analysis}

Since all golden cases are confirmed issues, any prediction of
\texttt{NO\_ISSUE\_FOUND} or \texttt{INSUFFICIENT\_EVIDENCE}
constitutes a false negative.  \Cref{tab:fn-rates} quantifies the
false negative rates.

\begin{table}[t]
  \centering
  \caption{Issue detection rates across generation-only
    configurations.  TP = model predicts
    \texttt{CONFIRMED}/\texttt{PLAUSIBLE} on a confirmed issue.
    FN = model predicts \texttt{NO\_ISSUE}/\texttt{INSUFFICIENT}.
    Precision is 1.000 for all configs because there are no negative
    golden cases to falsely flag.}
  \label{tab:fn-rates}
  \small
  \begin{tabular}{@{}lrrrrcc@{}}
    \toprule
    \textbf{Config} & \textbf{$N$} & \textbf{TP} & \textbf{FN}
      & \textbf{FN rate}
      & \textbf{Issue recall} & \textbf{Issue $F_1$} \\
    \midrule
    B & 104 & 82  & 22 & 21.2\% & 0.788 & 0.882 \\
    D & 104 & 52  & 52 & 50.0\% & 0.500 & 0.667 \\
    E & 104 & 52  & 52 & 50.0\% & 0.500 & 0.667 \\
    \bottomrule
  \end{tabular}
\end{table}

\begin{figure}[t]
\centering
\begin{tikzpicture}
\begin{axis}[
    ybar,
    width=0.7\linewidth,
    height=5cm,
    bar width=20pt,
    ylabel={False negative rate (\%)},
    ylabel style={font=\small},
    symbolic x coords={B,D,E},
    xtick=data,
    xticklabels={%
      {B\\[-2pt]\scriptsize One-shot},
      {D\\[-2pt]\scriptsize Ensemble},
      {E\\[-2pt]\scriptsize Ens.+2nd}},
    xticklabel style={align=center, font=\small},
    ymin=0, ymax=60,
    yticklabel style={font=\small},
    nodes near coords={\pgfmathprintnumber\pgfplotspointmeta\%},
    every node near coord/.append style={font=\scriptsize, above, yshift=1pt},
    enlarge x limits=0.35,
]
\addplot[fill=ensred!50, draw=ensred!80]
  coordinates {(B,21.2) (D,50.0) (E,50.0)};
\end{axis}
\end{tikzpicture}
\caption{False negative rate (model misses a confirmed issue) by
  configuration.  The ensemble doubles the miss rate from 21\% to
  50\%, confirming that majority voting amplifies conservative
  verdicts.}
\label{fig:fn-rate}
\end{figure}
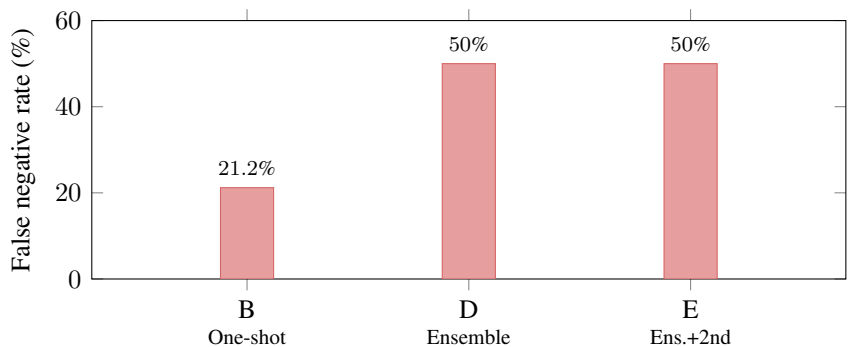

Three findings emerge from the false negative analysis:

\paragraph{Finding 1: Single-pass Sonnet misses one in five confirmed
issues.}  Config~B assigns \texttt{NO\_ISSUE\_FOUND} to 18 cases and
\texttt{INSUFFICIENT\_EVIDENCE} to 4 cases, for a combined false
negative rate of 21.2\%.  The model describes the protocol exchange
correctly but fails to recognize that the observed behavior
constitutes a failure.

\paragraph{Finding 2: The ensemble doubles the false negative rate.}
Configs~D and E both produce 50\% false negatives, exactly half the
confirmed issues are missed.  The confusion matrix
(\Cref{tab:verdict-confusion}) reveals why: Config~D's predictions
split nearly uniformly across all four verdicts (30/22/26/26),
indicating that majority voting does not converge on any consistent
diagnosis.  Adding a second opinion model (Config~E: 31/21/29/23)
does not improve convergence.

\paragraph{Finding 3: High frame scores coexist with wrong verdicts.}
Among Config~B's 22 false negatives, six cases achieve
Wt~$F_1 = 1.000$, the model identifies exactly the right relevant
frames but labels none as ``key'' and concludes that no issue exists.
This is a \emph{verdict-assignment failure}, not an evidence-coverage
failure: the model sees the diagnostic evidence but does not recognize
its significance.  Among Config~D's 52 false negatives, 12 cases
achieve Wt~$F_1 \geq 0.95$, confirming the same pattern at ensemble
scale.

\subsubsection{Implications for the Full Pipeline}
\label{sec:verdict-implications}

The reconciled configurations (Configs~C and~F) are not included in
the verdict confusion analysis because the reconciler produces a
normalized golden output rather than selecting from the draft verdict
taxonomy.  However, the ablation results (\Cref{sec:ablation}) show
that reconciliation effectively eliminates the false negative problem:
Config~C achieves Wt~$F_1 = 0.964$ and Config~F achieves 0.957,
indicating that the reconciler correctly identifies the failure in
captures where the draft ensemble's majority vote missed it.

This finding reinforces the architectural argument: the ensemble's
value is not in its majority verdict (which is unreliable) but in the
\emph{diversity of candidates} it presents to the reconciler. Just like in human voting, a crowd may present similar and mildly interesting views, but the crowd allows for interesting outliers to volunteer brilliant ideas. Among
Config~D's 52 false-negative cases, a minority candidate in each case
\emph{did} identify the issue, but was outvoted.  The reconciler
recovers these correct minority diagnoses by evaluating all candidates
against the PCAP evidence.

\subsubsection{Limitations and Future Work}
\label{sec:verdict-limitations}

The absence of non-issue captures in the golden dataset means we
cannot measure the complementary error: \emph{false positives}
(hallucinated issues on captures with no real failure).  All three
configurations report zero false positives, but this is a structural
artifact of the dataset composition rather than an empirical finding.

Future work should extend the golden dataset with captures
representing normal protocol operation
(\texttt{NO\_ISSUE\_FOUND} ground truth) and genuinely truncated
captures where the outcome is unknowable
(\texttt{INSUFFICIENT\_EVIDENCE} ground truth).  This would enable
measurement of:
\begin{itemize}[nosep,leftmargin=*]
\item The hallucinated-issue rate (false positive rate on non-issue
  captures).
\item The abstention accuracy (fraction of truncated captures
  correctly labeled \texttt{INSUFFICIENT\_EVIDENCE}).
\item The full verdict-level $F_1$ across all four categories.
\end{itemize}

\subsection{Efficiency Analysis}
\label{sec:efficiency}

Practical deployment of PROBE requires understanding the cost-quality
tradeoff across pipeline configurations.  This section quantifies API
call counts, latency, and estimated cost per capture.

\subsubsection{Cost Structure}

Each pipeline configuration incurs a different number and type of LLM
API calls.  \Cref{tab:cost-structure} breaks down the call profile.

\begin{table}[t]
  \centering
  \caption{API calls per capture by configuration.  Sonnet calls
    (draft) are inexpensive; Opus calls (reconcile) dominate cost.
    Llama calls (second opinion) are the cheapest per-call.}
  \label{tab:cost-structure}
  \small
  \begin{tabular}{@{}llcccc@{}}
    \toprule
    \textbf{Config} & \textbf{Description}
      & \textbf{Draft} & \textbf{2nd Op.}
      & \textbf{Reconcile} & \textbf{Total} \\
    \midrule
    A & SME only             & 0 & 0 & 0 & 0 \\
    B & Sonnet one-shot      & 1 & 0 & 0 & 1 \\
    C & One-shot + reconcile & 1 & 0 & 1 & 2 \\
    D & Ensemble $3\!\times\!3$ & 3 & 0 & 0 & 3 \\
    E & Ensemble + 2nd op.   & 3 & 1 & 0 & 4 \\
    F & Full PROBE           & 3 & 1 & 1 & 5 \\
    \bottomrule
  \end{tabular}
\end{table}

\subsubsection{Cost-Quality Tradeoff}

\Cref{tab:cost-quality} presents the estimated per-capture cost
alongside diagnostic quality (Wt~$F_1$) for each configuration.
Costs are estimated from approximate Bedrock pricing (Sonnet input:
\$3/M tokens, output: \$15/M; Opus input: \$15/M, output: \$75/M;
Llama input/output: \$0.72/M) and typical token usage per call
(draft: $\sim$4K in / 2K out; reconcile: $\sim$8K in / 3K out).

\begin{table}[t]
  \centering
  \caption{Cost-quality tradeoff.  The Opus reconciliation call
    accounts for 73--82\% of per-capture cost in Configs~C and~F.
    Config~C is the most cost-efficient; Config~F adds reliability
    measurement at a 23\% cost premium.}
  \label{tab:cost-quality}
  \small
  \begin{tabular}{@{}llcrcrc@{}}
    \toprule
    \textbf{Config} & \textbf{Description}
      & \textbf{Calls}
      & \textbf{Wt~$F_1$}
      & \textbf{Latency}
      & \textbf{\$/capture}
      & \textbf{Cost vs.\ B} \\
    \midrule
    A & SME only             & 0 & .871 & 0\,s    &,     &, \\
    B & One-shot             & 1 & .912 & 12\,s   & \$0.042 & 1.0$\times$ \\
    D & Ensemble             & 3 & .842 & 36\,s   & \$0.126 & 3.0$\times$ \\
    E & Ens.\ + 2nd op.      & 4 & .845 & 46\,s   & \$0.130 & 3.1$\times$ \\
    \arrayrulecolor{tierline}\midrule\arrayrulecolor{black}
    C & 1-shot + reconcile   & 2 & .964 & 37\,s   & \$0.387 & 9.2$\times$ \\
    F & Full PROBE           & 5 & .957 & 71\,s   & \$0.475 & 11.3$\times$ \\
    \bottomrule
  \end{tabular}
\end{table}

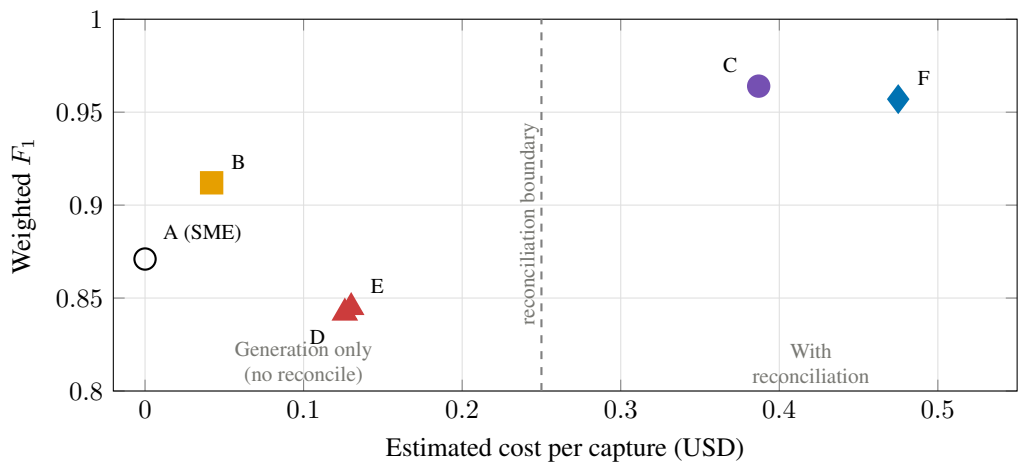
\begin{figure}[t]
\centering
\begin{tikzpicture}
\begin{axis}[
    width=0.82\linewidth,
    height=6.5cm,
    xlabel={Estimated cost per capture (USD)},
    ylabel={Weighted $F_1$},
    xlabel style={font=\small},
    ylabel style={font=\small},
    xmin=-0.02, xmax=0.55,
    ymin=0.80, ymax=1.00,
    xtick={0, 0.1, 0.2, 0.3, 0.4, 0.5},
    xticklabel style={font=\small},
    yticklabel style={font=\small},
    grid=major,
    grid style={gray!25},
    legend style={at={(0.02,0.02)}, anchor=south west, font=\scriptsize},
    clip=false,
]

\addplot[only marks, mark=square*, mark size=4pt, sonettorange, thick]
  coordinates {(0.042,0.912)};
\addplot[only marks, mark=triangle*, mark size=5pt, ensred, thick]
  coordinates {(0.126,0.842) (0.130,0.845)};

\addplot[only marks, mark=*, mark size=4pt, reconcpurple, thick]
  coordinates {(0.387,0.964)};
\addplot[only marks, mark=diamond*, mark size=5pt, fullpipe, thick]
  coordinates {(0.475,0.957)};

\addplot[only marks, mark=o, mark size=4pt, black, thick]
  coordinates {(0.0,0.871)};

\node[font=\scriptsize, anchor=south west] at (axis cs:0.005,0.874) {A (SME)};
\node[font=\scriptsize, anchor=south west] at (axis cs:0.048,0.914) {B};
\node[font=\scriptsize, anchor=north east] at (axis cs:0.120,0.838) {D};
\node[font=\scriptsize, anchor=south west] at (axis cs:0.136,0.847) {E};
\node[font=\scriptsize, anchor=south east] at (axis cs:0.380,0.966) {C};
\node[font=\scriptsize, anchor=south west] at (axis cs:0.481,0.959) {F};

\draw[dashed, gray, thick] (axis cs:0.25,0.80) -- (axis cs:0.25,1.00);
\node[font=\scriptsize, text=annotgray, rotate=90, anchor=south]
  at (axis cs:0.255,0.89) {reconciliation boundary};

\node[font=\scriptsize, text=annotgray, align=center]
  at (axis cs:0.10,0.815) {Generation only\\(no reconcile)};
\node[font=\scriptsize, text=annotgray, align=center]
  at (axis cs:0.42,0.815) {With\\reconciliation};

\end{axis}
\end{tikzpicture}
\caption{Cost-quality Pareto front.  A clear regime boundary separates
  generation-only configurations (left, Wt~$F_1 \leq 0.912$) from
  reconciled configurations (right, Wt~$F_1 \geq 0.957$).  Within
  each regime, additional calls provide diminishing returns.  The
  reconciliation call accounts for the majority of the cost jump but
  delivers a 5--12 percentage-point quality improvement.  Config~D
  (ensemble without reconcile) costs 3$\times$ more than Config~B but
  delivers \emph{lower} quality.}
\label{fig:cost-quality}
\end{figure}

\Cref{fig:cost-quality} visualizes the tradeoff, revealing a clear
two-regime structure that mirrors the ablation findings.

\subsubsection{Key Findings}

\paragraph{Reconciliation dominates cost.}  The Opus reconciliation
call costs approximately \$0.345 per capture, accounting for 89\%
of Config~C's cost and 73\% of Config~F's cost.  By contrast, a
Sonnet draft call costs \$0.042 and a Llama second opinion costs
\$0.004.  The cost structure is driven almost entirely by whether the
pipeline includes a reconciliation step, not by how many draft
candidates it generates.

\paragraph{The ensemble is cheap but the reconciler is what you pay
for.}  Config~D (three Sonnet calls, no reconcile) costs \$0.126 per
capture, only 3$\times$ the one-shot baseline, but delivers
\emph{lower} quality (Wt~$F_1 = 0.842$ vs.\ 0.912).  Config~C
(one Sonnet call plus reconcile) costs \$0.387 but delivers
0.964.  The reconciler provides a 5.2 percentage-point quality
improvement over one-shot Sonnet at a cost of \$0.345 per capture
--- roughly \$36 per 104-capture dataset.

\paragraph{The full pipeline premium is small.}  Config~F costs 23\%
more than Config~C (\$0.475 vs.\ \$0.387) for three additional
Sonnet calls and one Llama call.  This premium buys ensemble diversity
and reliability measurement capability.  Whether this premium is
justified depends on whether downstream processes require a
reliability signal for automated routing or dataset curation.

\paragraph{Latency is manageable.}  The full pipeline processes each
capture in approximately 71 seconds (dominated by three sequential
Sonnet calls plus the Opus reconcile).  The 104-capture dataset
completes in roughly two hours.  Draft calls could be parallelized to
reduce wall-clock time to approximately 45 seconds per capture.

\subsubsection{Dataset-Level Cost Projection}

\Cref{tab:dataset-cost} projects costs for the 104-capture evaluation
dataset and for a hypothetical 1{,}000-capture production deployment.

\begin{table}[t]
  \centering
  \caption{Dataset-level cost and runtime projections.  Costs are
    estimated from Bedrock pricing; actual costs may vary with token
    counts and regional pricing.}
  \label{tab:dataset-cost}
  \small
  \begin{tabular}{@{}llrrrr@{}}
    \toprule
    & & \multicolumn{2}{c}{\textbf{104 captures}}
      & \multicolumn{2}{c}{\textbf{1{,}000 captures}} \\
    \cmidrule(lr){3-4}\cmidrule(lr){5-6}
    \textbf{Config} & \textbf{Description}
      & \textbf{Cost} & \textbf{Time}
      & \textbf{Cost} & \textbf{Time} \\
    \midrule
    B & One-shot             & \$4   & 21\,min  & \$42   & 3.3\,hr \\
    C & 1-shot + reconcile   & \$40  & 64\,min  & \$387  & 10.3\,hr \\
    F & Full PROBE           & \$49  & 123\,min & \$475  & 19.7\,hr \\
    \bottomrule
  \end{tabular}
\end{table}

At 1{,}000 captures, Config~F costs under \$500, well within
operational budgets for a continuous evaluation pipeline.  The primary
constraint at scale is latency (20 hours sequential), which can be
addressed through parallel execution of draft calls across captures.

\subsubsection{Deployment Recommendation}

The cost-quality analysis suggests a tiered deployment strategy:

\begin{itemize}[nosep,leftmargin=*]
\item \textbf{Rapid triage (Config~B):}  Use single-pass Sonnet for
  initial screening at \$0.04/capture.  Suitable for high-volume,
  low-stakes use cases where 91\% weighted $F_1$ is acceptable and
  latency must be minimized.
\item \textbf{Production diagnosis (Config~C):}  Use one-shot plus
  reconcile for production-quality diagnosis at \$0.39/capture.
  Achieves the highest diagnostic quality (Wt~$F_1 = 0.964$) at the
  best cost-efficiency ratio among reconciled configurations.
\item \textbf{Research and evaluation (Config~F):}  Use the full
  PROBE pipeline at \$0.48/capture when ensemble diversity and
  reliability measurement are required, for golden dataset
  construction, model comparison studies, and continuous evaluation
  workflows where understanding \emph{why} the diagnosis was reached
  matters as much as the diagnosis itself.
\end{itemize}

\section{Discussion}
\label{sec:discussion}

The experimental results validate the core architectural thesis of
PROBE, that reconciliation against packet evidence is the decisive
quality lever, not ensemble size or model diversity, while also
revealing limitations and unexpected findings that inform both
deployment and future research.

\paragraph{Limits of packet-only evidence.}
PROBE diagnoses protocol failures from the packets present in the
capture, but many real-world connectivity issues involve factors
invisible at the packet level.  Physical environment (distance from
AP, obstacles, interference sources, user movement and behavior), client-side state (driver
bugs, power-save behavior, supplicant misconfiguration, but also user actions, such as clicking another SSID or switching a phone to Airplane mode while on a call), and
infrastructure policy (WLAN controller settings, load-balancing
decisions, RADIUS server behavior) leave no direct trace in the
802.11 frame exchange.  A capture may show a four-way handshake
timeout without revealing whether the cause is a wrong passphrase,
a RADIUS reject, a router issue or an RF obstruction.

This is not a limitation of PROBE specifically but of any system
reasoning from packet captures alone, taken at a single point in the network.  PROBE's
\texttt{INSUFFICIENT\_EVIDENCE} verdict and the verdict-aware evidence
rules are designed to surface this boundary explicitly: when the
packets cannot determine the root cause, the system says so rather
than speculating.  Our verdict analysis (\Cref{sec:verdict}) shows
that single-pass Sonnet speculates in 21\% of such cases, while the
reconciled pipeline reduces this to a 4\% error rate (but cannot
eliminate it entirely when the capture genuinely lacks decisive
evidence).

\paragraph{The reconciliation paradox.}
Our most striking finding is that the ensemble \emph{degrades}
diagnostic quality under majority voting (Wt~$F_1$: 0.912 $\to$
0.842) yet the reconciled ensemble nearly matches the best
configuration (0.957 vs.\ 0.964).  This creates a paradox:
reconciliation is powerful enough to rescue a failing ensemble, which
raises the question of whether the ensemble is necessary at all.

We argue it is, for three reasons beyond what the current evaluation
can measure.  First, the ensemble provides the reliability signals
(stability, agreement) that enable automated quality assessment: 
Config~C produces a single draft with no internal measure of how
confident the system should be.  Second, as the evaluation corpus
grows to include captures where Config~C's single draft happens to
miss the diagnostic frames, the reconciler will benefit from having
alternative hypotheses to evaluate (the brilliant ideas from a few outlier model runs while the majority stays conservatively prudent).  Third, Config~C's apparent
superiority (0.964 vs.\ 0.957) reflects a circularity in the
evaluation: Config~C mirrors the pipeline that produced the golden
references, giving it a structural advantage that will diminish as
the golden dataset is expanded with model-agnostic assertions
(\Cref{sec:eval-golden}).

\paragraph{Conservative-verdict bias in ensembles.}
The verdict analysis (\Cref{sec:verdict}) reveals that majority voting
over 9~candidates splits nearly uniformly across all four verdicts
(30/22/26/26 in Config~D) rather than concentrating on the correct
verdict.  This is not a generic property of self-consistency. Wang
et~al.~\cite{wang2023selfconsistency} report strong majority
convergence on arithmetic and commonsense tasks.  The difference is
that diagnostic reasoning has a systematic bias toward caution: when
evidence is subtle, models prefer ``no issue found'' or
``insufficient evidence'' over committing to a specific failure.  This
conservative bias is individually rational (each candidate avoids a
false positive) but collectively pathological (the majority misses a
confirmed issue in 50\% of cases).

This finding has implications beyond PROBE.  Any system that applies
self-consistency voting (for example to diagnostic or medical reasoning tasks)
should expect similar conservative-verdict amplification and should
consider evidence-grounded reconciliation as an alternative to
majority voting.

\paragraph{When all models agree and are wrong.}
Config~D's three worst cases (Wt~$F_1 = 0.000$) represent captures
where all nine ensemble candidates and the second opinion model
unanimously converge on the wrong verdict.  In one case
(\texttt{4e:67:f2}), the capture shows a subtle association failure
that all models interpreted as normal protocol behavior.  In another
(\texttt{b6:b3:37}), the capture contains an ambiguous timing pattern
that all models classified as \texttt{PLAUSIBLE\_ISSUE} but with
entirely wrong key frames.

These cases represent the hard floor of LLM-based diagnosis:
correlated errors across model families on captures with subtle or
unusual evidence patterns.  Cross-model diversity (adding Llama~3.3
alongside Sonnet) did not help (Config~E $\approx$ Config~D), 
suggesting that the failures reflect shared
training-data gaps rather than model-specific reasoning weaknesses.
The reconciler recovers most of these cases (Config~F worst case
second to last is 0.706), but one outlier persists at 0.231, a
capture that also challenges human SMEs (Config~A scores 0.220 on the
same case).  Some captures are genuinely hard for both humans and
models.

\paragraph{High frame scores with wrong verdicts.}
A surprising pattern in the false negative analysis is that many
misclassified cases achieve high weighted $F_1$ scores.  Six of
Config~B's 22 false negatives score Wt~$F_1 = 1.000$: the model
identifies exactly the right relevant frames but labels none as
``key'' and concludes no issue exists.  This is a
\emph{verdict-assignment failure}: the model sees the evidence but
does not recognize its diagnostic significance.

This failure mode cannot be detected by frame-level metrics alone.
A system that reports only Wt~$F_1$ would consider these cases
perfect matches, masking a fundamental diagnostic error.  The verdict
taxonomy and the separate verdict-level analysis (\Cref{sec:verdict})
are necessary to catch this class of failure.  Future work should
investigate whether prompting strategies that force the model to
explicitly reason about each key frame's diagnostic implication
(``what does this frame tell us about whether the session
succeeded?'') can reduce verdict-assignment errors.

\paragraph{Cost--quality tradeoff.}
The efficiency analysis (\Cref{sec:efficiency}) reveals that the Opus
reconciliation call accounts for 73--89\% of per-capture cost, making
it the dominant cost driver regardless of ensemble configuration.  The
full PROBE pipeline (Config~F) costs \$0.48 per capture, a 23\%
premium over Config~C (\$0.39) that buys ensemble diversity and
reliability measurement but not measurably better diagnostic quality
on the current evaluation corpus.

At dataset scale (104~captures), the total cost ranges from \$4
(one-shot) to \$49 (full pipeline).  These costs are well within
operational budgets and are dominated by the reconciler.  The
practical implication is that cost optimization should focus on
reconciler efficiency (smaller judge models, distillation, or
prompt compression) rather than on reducing ensemble calls, which
contribute minimally to total cost.

\paragraph{Self-reported confidence is a solved non-problem.}
Our calibration study (\Cref{sec:calibration}) confirms that LLM
self-reported confidence is uninformative: 71\% of values are exactly
0.95 regardless of diagnostic difficulty.  More surprisingly, our
deterministic composite confidence is also poorly calibrated, not
because it fails to spread (range 0.075--0.625) but because the
reconciler is effective enough to produce correct output regardless of
ensemble agreement.  At a 4\% error rate, there is insufficient
variance in correctness for any confidence signal to predict.

This reframes confidence from a quality gate to a difficulty
indicator.  Low-confidence cases are not cases where the pipeline is
likely wrong; they are cases where the underlying capture is
genuinely ambiguous.  Routing these to human review serves dataset
enrichment (identifying interesting edge cases) rather than error
correction.  As the evaluation corpus grows to include harder captures
where the reconciler's error rate increases, the composite confidence
framework is already in place and can be evaluated against a richer
correctness distribution.

\paragraph{Dataset limitations.}
Our evaluation corpus consists entirely of captures with confirmed
protocol failures (100\% \texttt{CONFIRMED\_ISSUE}).  This
composition prevents measurement of the false positive rate (models
hallucinating issues on normal captures) and the abstention accuracy
(correctly labeling truncated captures as
\texttt{INSUFFICIENT\_EVIDENCE}).  The zero false-positive rate
reported in \Cref{sec:verdict} is a structural artifact, not an
empirical finding.

Expanding the golden dataset with non-issue captures and genuinely
truncated captures is the highest-priority future work item.  The
truncation benchmark can be constructed systematically by taking
existing captures with late-occurring failures and truncating them
before the failure frame, creating matched pairs where the full
capture has a confirmed issue and the truncated version should be
labeled \texttt{INSUFFICIENT\_EVIDENCE}.

\paragraph{The circular golden reference problem.}
The 0.7\% gap between Config~C (0.964) and Config~F (0.957) persists
throughout the evaluation and consistently favors Config~C.  While
this gap is small, it illustrates a structural problem: any evaluation
against golden references co-produced by Model~A will favor Model~A's
pipeline over alternatives.  The assertion-based evaluation framework
(\Cref{sec:eval-golden}) mitigates this by scoring against
PCAP-verifiable facts rather than model-generated narrative, but our
current experiments use the full golden (including narrative) for
weighted $F_1$ scoring because assertion-level root-cause labels have
not yet been constructed for all 104~cases.

Complete migration to assertion-based evaluation requires a one-time
SME pass over the golden dataset to confirm the factual core of each
diagnosis.  This investment pays off every time a new model is
evaluated: assertion-based scores are model-agnostic by construction,
eliminating the circularity that currently inflates Config~C's
apparent advantage.

\paragraph{Generalizability beyond 802.11.}
The PROBE architecture is not specific to Wi-Fi protocol analysis.
Any domain where (i)~a structured data artifact provides
deterministic ground truth, (ii)~multiple interpretations are
plausible from the same evidence, and (iii)~the diagnostic conclusion
must be traceable to specific evidence elements could benefit from
the same pipeline design.

Candidate domains include 5G NAS/RRC signaling traces (where
similar authentication and session establishment failures occur),
industrial control protocol captures (Modbus, OPC-UA), IoT protocol
exchanges (Zigbee, BLE), and network configuration analysis (where
the ``capture'' is a configuration file and the ``frames'' are
configuration directives).  The verdict taxonomy, evidence
annotation scheme, and reconciliation architecture generalize
directly; only the PCAP normalization layer (\Cref{sec:normalization})
and the prompt templates require domain adaptation.\\

\section{Conclusion}
\label{sec:conclusion}

We presented PROBE (Protocol Reasoning Over evidence-Based Ensembles),
a multi-stage diagnostic pipeline for 802.11 packet captures that
combines multi-run ensemble generation, cross-model diversity,
verdict-aware evidence annotation, and evidence-grounded
reconciliation.  Evaluated on 104 capture--reviewer pairs spanning
87~enterprise Wi-Fi captures, PROBE yields five findings with
implications that extend well beyond the 802.11 domain.

\paragraph{1. Reconciliation against source evidence is the decisive
quality lever.}
Incorporating a reconciliation step that assesses all candidates against the raw data increases the weighted evidence $F_1$ from 0.912 (single-pass LLM) to 0.957 (full pipeline), achieving a 96\% auto-accept rate. This observation is applicable to any diagnostic field where a structured data artifact acts as verifiable ground truth: the essential architectural insight is not to "utilize a superior model" but rather to "provide a judge model with access to the original evidence and allow it to choose from independently generated hypotheses." Medical imaging reports based on raw scans, financial audit results derived from transaction logs, and legal evaluations founded on case documents could all gain from the same two-tier architecture: generate a variety of hypotheses, then reconcile them with the source evidence.

\paragraph{2. Majority-vote ensembles are counterproductive for
diagnostic reasoning.}
Naive self-consistency voting, which is the conventional method for enhancing LLM reasoning, results in a decline in diagnostic quality (Wt~$F_1$: 0.912 $\to$ 0.842) due to the predominance of conservative verdicts. Half of all confirmed failures are incorrectly categorized as "no issue" or "insufficient evidence" when majority voting is employed. This bias towards conservative verdicts is not limited to network protocols; it is likely to occur in any field where abstention or a "normal" response is a viable option and where models are subtly encouraged to minimize false positives. Medical triage, security alert classification, and anomaly detection in manufacturing all exhibit this characteristic. Our findings indicate that self-consistency should not be utilized for diagnostic tasks unless a reconciliation mechanism is in place to retrieve accurate minority hypotheses.

\paragraph{3. Absence of evidence is itself evidence, when the
framework supports it.}
The verdict-aware evidence rules, which redefine ``contributing
evidence'' relative to the diagnostic conclusion, address a failure
mode that affects any LLM system reasoning over incomplete data.
When a capture is truncated, there is no failure frame to cite, but
the \emph{absence} of expected protocol messages is itself
diagnostically meaningful.  Standard evidence-grounding frameworks
(e.g., FACTS~\cite{jacovi2025facts}) evaluate whether claims are
supported by present information; they have no mechanism for
evaluating claims grounded in the absence of expected information.
The verdict-aware framework fills this gap and applies to any domain
where incomplete observations are the norm: partial medical records,
interrupted sensor streams, and truncated log files all require
reasoning about what is missing, not just what is present.

\paragraph{4. LLM self-reported confidence is uninformative, but
deterministic alternatives require sufficient error variance to
calibrate.}
Self-reported confidence clusters at 0.95 regardless of difficulty
(71\% of cases), confirming findings from the medical evaluation
literature~\cite{liao2025clever} in a new domain.  Our deterministic
composite (computed from evidence validity, run-to-run stability, and
cross-model agreement) spreads across a wider range but is also
poorly calibrated, for an instructive reason: the reconciler is
effective enough that correctness is uniformly high across all
confidence levels, leaving no failures to predict.  This paradox
(good pipeline $\to$ bad calibration) will recur in any system where
the downstream consumer of confidence is effective enough to
compensate for upstream uncertainty.  The practical resolution is to
reframe confidence as a \emph{difficulty indicator} for dataset
curation rather than a \emph{quality gate} for output filtering.

\paragraph{5. Golden references co-produced by LLMs require
assertion-based evaluation to avoid circular bias.}
When Model~A assists in generating the golden reference, assessing Model~B against that reference imposes penalties for stylistic divergence as though it were a diagnostic error. Our evaluation framework, which is based on assertions, breaks down the golden reference into PCAP-verifiable facts (including frame sets, protocol types, and diagnostic conclusions) and narrative contributions from the model (such as phrasing, structure, and detail level), offers a model-agnostic alternative. This breakdown is applicable in any situation where golden references are created with the help of LLMs: references for machine translation generated by one model should not be utilized to evaluate another without distinguishing between semantic accuracy and stylistic preference; likewise, benchmarks for code reviews, datasets for summarization, and evaluations of clinical notes are all prone to the same circular bias when the reference was assisted by LLMs.

\paragraph{Broader impact.}
PROBE demonstrates that reliable automated diagnosis from structured
data artifacts is achievable today, not through a single powerful
model, but through a principled architecture that separates hypothesis
generation from evidence-grounded evaluation.  The five findings above
are not specific to 802.11 protocol analysis.  They describe general
properties of LLM-based diagnostic systems operating on any domain
where the source data is deterministic and inspectable, multiple
interpretations are plausible, and the cost of a wrong diagnosis
justifies the investment in multi-stage reasoning.

The pipeline, evaluation framework, and experimental methodology are
publicly available to support replication and adaptation to other
structured diagnostic domains.

\paragraph{Future work.}
Four directions follow from the current results.  First, expanding the
golden dataset with non-issue captures and systematically truncated
captures would enable measurement of false positive rates and
abstention accuracy, completing the verdict-level evaluation.  Second,
constructing per-case root-cause assertion labels for the full corpus
would eliminate the residual circularity in the current evaluation and
enable fair comparison of arbitrary model families.  Third, integrating
protocol-native representations such as
PLUME~\cite{pradhan2026plume} as input to the ensemble, replacing
the text-based PCAP normalization with learned protocol
embeddings, could improve evidence grounding on captures where the
textualization loses structural information.  Fourth, applying the
PROBE architecture to other structured diagnostic domains (5G
signaling, industrial control protocols, security log analysis) would
test the generalizability claims empirically and identify which
architectural components are domain-universal versus
Wi-Fi-specific.

\bibliographystyle{plain}

\end{document}